\documentclass[10pt,twocolumn,letterpaper]{article}

\usepackage{cvpr}
\usepackage{graphicx}
\usepackage{amsmath,amssymb}
\usepackage{booktabs}
\usepackage{array}
\usepackage{multirow}
\usepackage{xcolor}
\usepackage{enumitem}
\usepackage{url}
\usepackage{pifont}

\newcommand{\method}{EmoScope}
\definecolor{oursaccent}{RGB}{125,45,175}
\newcommand{\cmark}{{\color{green!50!black}\checkmark}}
\newcommand{\xmark}{{\color{red!70!black}\ensuremath{\boldsymbol\times}}}

\newcommand{\nFullEmoeditImg}{415}
\newcommand{\nFullEmoeditPairs}{3,320}
\newcommand{\nFullEmoeditorImg}{504}

\newcommand{\nFullArtemisImg}{34}

\newcommand{\nFullBenchPairs}{7,352}
\newcommand{\nHumanImages}{114}
\newcommand{\nHumanTasks}{912}
\newcommand{\nHumanQuestions}{1,824}
\newcommand{\nDesignedResponses}{5,472}
\newcommand{\nSessions}{299}
\newcommand{\nUniqueParticipants}{252}
\newcommand{\nAnswers}{4,693}
\newcommand{\nRawAnswers}{5,578}
\newcommand{\nRawParticipants}{391}
\newcommand{\pctFilteredOut}{16\%}

\newcommand{\qOneOverall}{88.1\%}

\newcommand{\qOneAngerVsEditor}{95.3\%}

\newcommand{\qOneAweVsEditor}{86.9\%}

\newcommand{\qOneContentmentVsEditor}{95.9\%}

\newcommand{\qOneExcitementVsEditor}{98.0\%}

\newcommand{\coherentOurs}{92.0\%}
\newcommand{\caseAcount}{124}
\newcommand{\caseBcount}{15}

\newcommand{\nLinkedPairs}{1,697}
\newcommand{\maxPearsonSpearmanDiff}{0.067}
\newcommand{\qOneVsEmoeditCI}{[81.2\%, 84.9\%]}
\newcommand{\qOneVsEmoeditorCI}{[91.8\%, 94.2\%]}
\newcommand{\nDemoMinPref}{73.5\%}
\newcommand{\nDemoMinSubgroup}{45+ (Age range)}
\newcommand{\nDemoMinN}{200}
\newcommand{\nDemoMaxPref}{94.2\%}
\newcommand{\nDemoMaxSubgroup}{35--44 (Age range)}
\newcommand{\nDemoNDims}{5}
\newcommand{\hardestEmo}{amusement}
\newcommand{\hardestEmoRej}{9.07\%}
\newcommand{\easiestEmo}{contentment}
\newcommand{\easiestEmoRej}{1.46\%}
\newcommand{\diffRatio}{6.2\,\times}
\newcommand{\nTotalTraces}{7,671}

\newcommand{\nHardestImages}{16}

\newcommand{\nInteractiveSessions}{96}
\newcommand{\nInteractiveParticipants}{10}

\newcommand{\pctInteractiveAccepted}{100\%}
\newcommand{\interactiveSatMean}{3.92}

\newcommand{\pctInteractiveSatHigh}{72\%}
\newcommand{\interactiveDurMedian}{161}

\newcommand{\interactivePlanIterMean}{1.72}
\newcommand{\interactiveEditIterMean}{1.67}

\newcommand{\pctInteractiveFeedback}{39\%}

\newcommand{\nWebuiRespondents}{10}

\newcommand{\webuiQQuality}{3.50}

\newcommand{\webuiQInspiring}{3.90}
\newcommand{\webuiQUnderstand}{3.80}

\newcommand{\webuiDialogAnyNecessaryPct}{100\%}
\newcommand{\webuiDialogVeryNecessaryN}{6}

\newcommand{\bestMetricR}{+0.51}

\newcommand{\worstMetricR}{-0.11}

\definecolor{cvprblue}{rgb}{0.21,0.49,0.74}
\usepackage[breaklinks,colorlinks,allcolors=cvprblue]{hyperref}

\def\confName{CVPR}
\def\confYear{2026}

\title{What Can I Edit? Open-Ended Strategy Discovery and the Emotion Editability Landscape}

\author{
Qing Li$^{1\dagger}$ \quad
Zeyu Dong$^{1\dagger}$ \quad
Yin Cui$^{2}$ \quad
Chuan Yan$^{3\ast}$ \quad
Xiaojiang Peng$^{1\ast}$\\
$^1$School of Artificial Intelligence, Shenzhen Technology University\\
$^2$School of Innovation Design, Shenzhen Technology University\\
$^3$Stanford University\\
{\tt\small \{liqing,zeyudong,pengxiaojiang\}@sztu.edu.cn, cuiyin@sztu.edu.cn, chuanyan@stanford.edu}\\
{\small $^\dagger$Equal contribution. $^\ast$Corresponding authors.}
}

\begin{document}
\maketitle

\begin{abstract}
Emotional image editing requires more than applying affective filters or modifying predefined visual factors: an effective edit must discover what a particular image can afford for a target emotion.
Existing affective image manipulation (AIM) methods, including recent agentic variants, still largely operate within bounded strategy spaces---retrieving, selecting, or refining edits from predefined factor taxonomies, knowledge libraries, or conventional editing templates.
As a result, they improve execution control but often miss image-specific, context-grounded strategies that humans immediately recognize as emotionally meaningful.
We introduce \method{}, a multi-agent framework that reframes the problem from ``how should I edit?'' to ``what can I edit?''
\method{} first discovers an image-specific editable space through emotion-conditioned affordance reasoning, then uses a semantic hierarchy of anchors, variables, and context to navigate the consistency--expressiveness tradeoff before executing and verifying the edit.
Because the plan is expressed as image-specific affordances rather than retrieved templates, this design also naturally exposes the editing strategy as an interactive surface, letting users refine emotional expression at the plan level rather than at the pixel level.
Through a large-scale human evaluation across all eight Mikels emotion categories (\nAnswers{} valid responses across \nHumanQuestions{} pairwise questions), humans strongly prefer \method{} over two competitive baselines (\qOneOverall{} mean preference); attribution data further shows that \method{} selects target-emotion-adaptive strategies rather than applying a uniform template.
The same affordance-level plan also supports lightweight user refinement, examined in a small interactive pilot.
We additionally show that classifier-based metrics (Emo-A, Emo-S) exhibit emotion-conditional blind spots toward non-stereotypical, context-grounded edits, and we present a relative content--emotion preference-affinity landscape revealing that \method's advantage varies systematically across image--emotion combinations.
\end{abstract}


\section{Introduction}
\label{sec:introduction}

Image editing has evolved from well-defined operations---adjusting brightness, applying filters, removing blemishes---to tasks driven by abstract human intentions.
A request like ``make this image feel happier'' has no single correct answer; it requires understanding the image's visual content, the cultural connotations of its elements, and the multifaceted nature of emotional expression.
This class of tasks, broadly termed \emph{affective image manipulation} (AIM)~\cite{yang2024emoedit}, sits at the intersection of image editing and affective computing, demanding not just technical proficiency but visual-emotional reasoning over an open-ended space of possible edits.

\begin{figure}[t]
\centering

\newcommand{\oursbox}[1]{%
  {\setlength{\fboxrule}{1.5pt}\setlength{\fboxsep}{0pt}%
   \fcolorbox{oursaccent}{white}{#1}}%
}
\newcommand{\baselinebox}[1]{%
  {\setlength{\fboxrule}{0.6pt}\setlength{\fboxsep}{0pt}%
   \fcolorbox{black!55}{white}{#1}}%
}

\centering

{\renewcommand{\arraystretch}{0}\setlength{\tabcolsep}{2pt}%
\begin{tabular}{@{}c@{\hspace{3pt}}c@{\hspace{3pt}}c@{}}
  {\footnotesize\scshape Source} &
  \multicolumn{2}{c}{\footnotesize\scshape Baselines (Amusement)} \\[2pt]
  \includegraphics[width=0.30\linewidth]{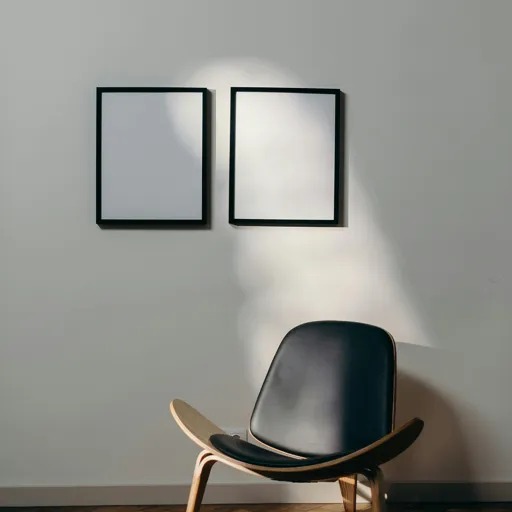} &
  \baselinebox{\includegraphics[width=0.30\linewidth]{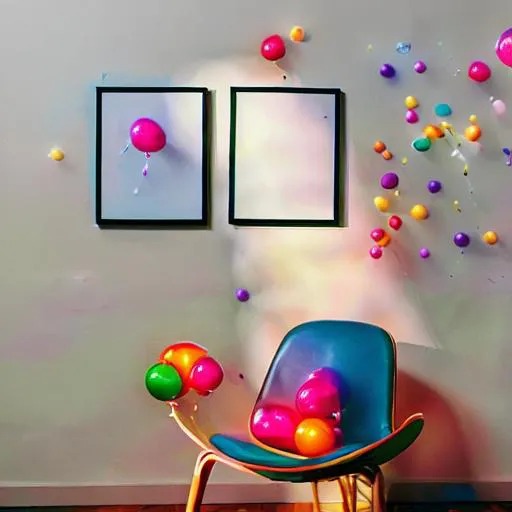}} &
  \baselinebox{\includegraphics[width=0.30\linewidth]{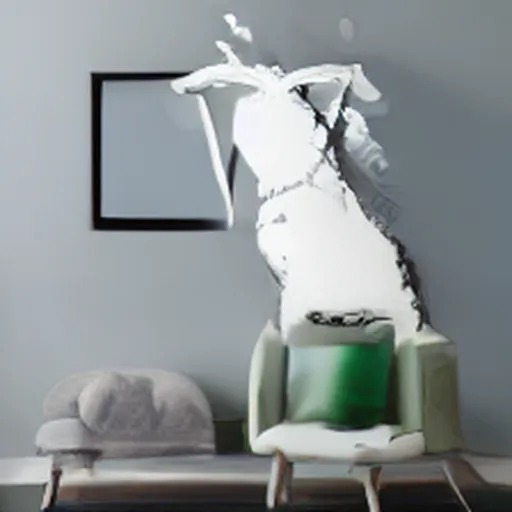}} \\[2pt]
   &
  {\scriptsize\textbf{EmoEdit}~{\itshape\color{black!60}\scriptsize CVPR'24}} &
  {\scriptsize\textbf{EmoEditor}~{\itshape\color{black!60}\scriptsize ICCV'25}} \\[1pt]
  & {\scriptsize\itshape scattered balls} & {\scriptsize\itshape lost subject} \\[6pt]
  \multicolumn{3}{@{}l@{}}{\footnotesize\color{oursaccent}\scshape Ours\, --\, same source, three emotions} \\[2pt]
  \oursbox{\includegraphics[width=0.30\linewidth]{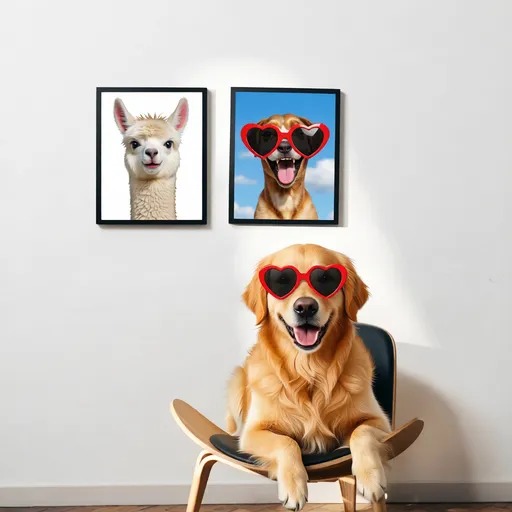}} &
  \oursbox{\includegraphics[width=0.30\linewidth]{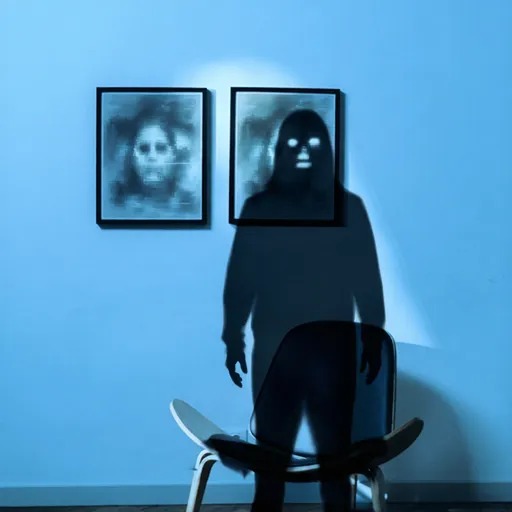}} &
  \oursbox{\includegraphics[width=0.30\linewidth]{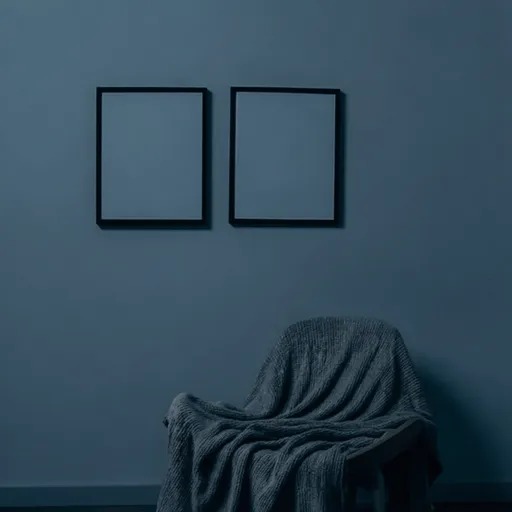}} \\[2pt]
  {\scriptsize\color{oursaccent}\textbf{Ours $\cdot$} \emph{Amusement}} &
  {\scriptsize\color{oursaccent}\textbf{Ours $\cdot$} \emph{Fear}} &
  {\scriptsize\color{oursaccent}\textbf{Ours $\cdot$} \emph{Sadness}} \\[1pt]
  {\scriptsize\itshape playful portraits} &
  {\scriptsize\itshape cursed portraits} &
  {\scriptsize\itshape stays empty} \\
\end{tabular}}

\caption{\textbf{A paradigm shift in emotional image editing.}
\emph{Top:} on \emph{Amusement}, EmoEdit~\cite{yang2024emoedit} produces image-agnostic stereotypes (scattered balls from a predefined factor taxonomy) while EmoEditor~\cite{ling2024emoeditor} pursues the emotion so aggressively that the subject is lost.
\emph{Bottom:} \method{} reframes the problem as \emph{open-ended discovery in a per-image editable space}: it identifies the two empty picture frames as themselves a canvas, then reuses the same affordance differently for each target---playful portraits for \emph{Amusement}, cursed portraits for \emph{Fear}, and deliberately empty frames for \emph{Sadness}---showing that the strategy is a function of image content rather than a selection from an enumerable menu.}
\label{fig:teaser}
\end{figure}

EmoEdit~\cite{yang2024emoedit} constructs a hierarchical emotion factor tree decomposing affect into object, scene, action, and facial expression factors.
EmoEditor~\cite{ling2024emoeditor} separates global context from local semantic regions via a dual-branch architecture.
These approaches differ in mechanism---factor trees, dual branches, noise-level mappings, knowledge graphs, even test-time optimization~\cite{muse2025}---but share a common property: the \emph{source of editing strategies} is bounded by each method's representational substrate, whether an explicit taxonomy or an implicit parameterization.
A more recent generation of agentic AIM systems---EmoAgent~\cite{emoagent2025}, EmoFeedback$^2$~\cite{emofeedback2025}, Emotion-Director~\cite{emotiondirector2025}---adds planning, critique, and feedback loops, contributing significant gains in controllability and error correction; the strategies on which these loops operate, however, are still drawn from a pre-existing source (knowledge library, factor templates, or conventional editing instructions), so the loops sharpen execution within an inherited strategy space rather than enlarging it.

While such bounded strategy spaces enable strong execution control within predefined boundaries, they leave a substantial unexplored territory of context-grounded strategies---the editable space of each image is image-specific and far larger than any predefined factor taxonomy can enumerate.
Figure~{1} makes this concrete.
Given a minimalist room with two empty picture frames and a single chair (hardly a typical ``amusement'' scene), EmoEdit applies its factor-tree taxonomy and scatters colored balls across the chair and walls, while EmoEditor pursues an amusement-classifier signal so aggressively that it loses the subject entirely.
\method{}, in contrast, identifies that the empty frames \emph{are themselves a canvas} and fills them with playful animal portraits paired with a matching subject in the chair---a strategy outside the vocabulary of any factor tree, requiring an understanding of what this specific image affords.
The same affordance is then used differently for other emotion targets: cursed portraits for \emph{Fear}, and---most tellingly---\emph{deliberately empty frames} for \emph{Sadness}, where the system treats absence itself as the expressive choice.
What unites these strategies across emotions is a single principle: they emerge from per-image affordance reasoning rather than retrieval, refinement, or selection within a pre-existing strategy space.
Crucially, a plan--verify loop alone does not necessarily solve this failure mode if the planner still chooses among stereotyped strategy sources---it requires changing what the planner does, not just adding verification on top.

This observation motivates a paradigm shift in how we approach emotional image editing.
Instead of asking \emph{``how should I edit this image?''} (selecting from predetermined strategies), we first ask \emph{``what can I edit in this image?''}---discovering the image-specific \emph{editable space} through emotion-conditioned reasoning about the image's visual elements.
In this view, the planner's role is not to choose the best operation from a known menu, but to \emph{construct the menu} from the image's affordances under a target emotion.
Once the editable space is identified, a vision-language model (VLM) can generate creative, contextually grounded editing strategies within it, unconstrained by a fixed strategy source.
While the idea of discovering editing possibilities has been explored in general creative editing~\cite{creativityvla2024,coeditorpp2026}, to our knowledge we are the first to apply it to emotion-specific editing, where the target emotion fundamentally reshapes \emph{which} elements are editable and \emph{how}.

Formalizing this view, we identify two orthogonal requirements for a successful emotional edit.
\textbf{Consistency}: the edited image must remain recognizable as originating from the original.
Critically, consistency is not monolithic---it operates at multiple levels from pixels to narrative, and relaxing it at higher levels dramatically expands the strategy space.
This creates a \emph{consistency--expressiveness tradeoff} that prior work has not explicitly modeled: existing agentic editors verify whether an edit succeeds, but rarely formalize \emph{which} kinds of consistency may be relaxed to make a more expressive emotional edit possible.
\textbf{Plausibility}: the edit must genuinely convey the intended emotion without being so extreme as to undermine its own message.
Together, these two constraints define the true editable space---image-specific, emotion-dependent, and far richer than prior work assumes.

Moreover, we raise a question that the field has only partially addressed: \emph{is every image equally amenable to every target emotion?}
AIM-Bench~\cite{aimbench2026} observes that positive emotions are generally more editable than negative ones at the \emph{model level}, but the variation at the \emph{individual image level}---which specific visual properties make an image more or less amenable to a particular emotional transformation---remains unquantified.
Intuitively, some (image content, target emotion) combinations have natural \emph{affinity} while others present fundamental \emph{resistance}.
As an empirical proxy for this larger question, we analyze how a method's \emph{relative} human-preference advantage varies across content type and target emotion, providing the first image-level map of where one method's gains over another concentrate or fade.

We present \textbf{\method{}}, a multi-agent framework that operationalizes these insights (Fig.~{1}).
Unlike other agentic AIM systems whose planner retrieves from a library or instantiates templates, \method's \textbf{Planner} generates editing strategies \emph{ab initio} from per-image affordances under a target emotion.
A \textbf{Plan Verifier}, \textbf{Image Editor}, and \textbf{Result Verifier} then form two iterative refinement loops that navigate the consistency--expressiveness tradeoff at runtime by falling back from ambitious strategies to safer ones when needed.
In this design, verification serves a different role than in earlier agentic AIM: not to compensate for a weak strategy source, but to keep open-ended discovery grounded, plausible, and consistent.


Our contributions are:

\begin{enumerate}[leftmargin=*,nosep]
    \item We formalize the \emph{consistency--expressiveness tradeoff} for emotional image editing and argue that prior methods---including recent agentic AIM systems---operate within bounded strategy sources that leave the broader space of context-grounded strategies unexplored.
    \item We propose \method{}, an agentic framework whose key contribution is an \emph{affordance-discovery Planner} that constructs the editable space \emph{ab initio} from per-image elements (not the plan--verify loop itself, now standard in image-editing agents), plus a semantic hierarchy that operationalizes the consistency--expressiveness tradeoff at runtime; because the plan is expressed at the affordance level, it also serves as an interactive surface for non-experts to refine edits at the plan level rather than the pixel level (pilot study in Sec.~{4.8}).
    \item We present the first image-level analysis of a \emph{relative content--emotion preference-affinity landscape}, quantifying how \method's human-preference advantage over baselines varies across image content and target emotion---complementing model-level observations~\cite{aimbench2026} and static emotion--scene correlations~\cite{emoscene2026}.
    \item Through a large-scale human evaluation ($\nAnswers{}$ valid responses across $\nHumanQuestions{}$ pairwise questions), we demonstrate that (a)~humans significantly prefer our edits, (b)~our method selects target-emotion-adaptive editing strategies whose distribution shifts predictably with the target emotion, and (c)~existing automated metrics (Emo-A, Emo-S) exhibit an emotion-conditional blind spot toward non-stereotypical, context-grounded edits.
\end{enumerate}


\section{Related Work}
\label{sec:related_work}

\subsection{Affective Image Manipulation}

Affective image manipulation (AIM) aims to edit images so that they evoke a specified target emotion.
Early approaches operated at individual visual layers: Affective Image Filter~\cite{weng2023aif} adjusts color tones, while C2A2~\cite{c2a2_2024} modifies compound facial expressions.
Recognizing that emotion is conveyed through multiple factors simultaneously~\cite{rao2016learning}, recent methods attempt multi-level editing.
EmoEdit~\cite{yang2024emoedit} constructs a hierarchical emotion factor tree (object, scene, action, facial expression) and edits each factor independently.
EmoEditor~\cite{ling2024emoeditor} employs a dual-branch architecture integrating global context with local semantic regions.
AIEdiT~\cite{aiedit2025} maps emotional factors to diffusion noise levels for coarse-to-fine editing.
EmoKGEdit~\cite{emokgedit2026} uses a knowledge graph to guide object-level adjustments followed by global atmospheric harmonization.

These methods differ substantially in their editing mechanisms---factor trees, parallel branches, noise-level mappings, knowledge graphs---but they share a common property: the \emph{vocabulary of editing strategies} is bounded by the method's representational substrate (e.g., EmoEdit's strategies are drawn from EmoSet~\cite{emoset2023} clusters, while continuous approaches like MUSE~\cite{muse2025} and AIEdiT's noise-level mapping define the space of possible edits implicitly through their parameterization).
Our work asks whether this bounded strategy space is sufficient, and demonstrates empirically that it is not.

Concurrent works explore complementary dimensions: generate-from-scratch (EmoGen~\cite{emogen2024}, EmotiCrafter~\cite{emoticrafterr2025}), polarity-decomposed (CoEmoGen~\cite{coemogen2025}), and style-based (EmoStyle~\cite{emostyle2025}) lines target different tasks than in-place emotional editing.
Regressor-Guided Emotion Editing~\cite{gebhardt2025} finds that diffusion-based methods enable semantic-level changes that optimization-based methods cannot, consistent with our thesis that higher-level strategies produce qualitatively different emotional effects.
AIM-Bench~\cite{aimbench2026} provides the first AIM benchmark and observes a positivity bias, hinting at the emotion-dependent editability variation we systematically investigate.

\subsection{Agent-Based and Creative Image Editing}

The plan--edit--verify paradigm for complex image editing inherits its design from the broader literature on language-model agents---ReAct~\cite{yao2023react} for interleaving reasoning with tool actions, and AutoGen~\cite{wu2024autogen} for orchestrating multi-agent conversations---specialised here to vision-language models.
Within the image-editing setting, GenArtist~\cite{genartist2024} decomposes goals into sub-tasks with step-by-step verification.
GraPE~\cite{grape2024} and IMAGAgent~\cite{imagagent2026} implement plan-execute-reflect loops.
MIRA~\cite{mira2025} uses iterative perception-reasoning-action cycles.
These architectures are now well-established; we adopt the dual-loop design not as a contribution in itself but as the infrastructure needed to support open-ended strategy exploration.

Within AIM, EmoAgent~\cite{emoagent2025} is the first multi-agent framework, employing Planning, Editing, and Critic agents.
While its Planning Agent performs top-$k$ retrieval from a fixed Emotion-Factor Knowledge library---claiming to move ``beyond fixed mappings''---the strategy vocabulary remains bounded by what has been observed in EmoSet.
EmoFeedback$^2$~\cite{emofeedback2025} introduces LVLM-based feedback for iterative refinement, and Emotion-Director~\cite{emotiondirector2025} mitigates affective shortcuts via chain-of-concept reasoning.
Both operate within conventional editing strategies.
Our Planner differs in that it generates strategies \emph{ab initio} through emotion-conditioned reasoning about visual elements: instead of retrieving from a library, it identifies what the image's elements \emph{can become} in the context of a target emotion.

In broader creative editing, Creativity-VLA~\cite{creativityvla2024} and CoEditor++~\cite{coeditorpp2026} address general creative editing from vague hints or "what vs. how" decomposition, and Beyond Pixels~\cite{beyondpixels2026} transfers creative essence from a reference image via schema-driven reasoning.
Our planner differs in being emotion-conditioned and reference-free, yielding emotion-specific strategies rather than general creative suggestions.

\subsection{Emotion Editability and Evaluation}

Automated AIM evaluation relies on emotion classifiers: Emo-A (classification accuracy) and Emo-S (confidence) measure whether edited images are classified as the target emotion~\cite{emoagent2025}.
LPIPS~\cite{zhang2018unreasonable} and CLIP-I~\cite{radford2021clip} quantify perceptual and semantic distance, complementing the older pixel-level family of structural-similarity (SSIM~\cite{wang2004ssim}) and Fr{\'e}chet Inception Distance (FID~\cite{heusel2017fid}) measures used in image generation more broadly.
These classifier-based metrics, trained on images where emotion is conveyed through surface-level features~\cite{rao2016learning}, may fail to evaluate emotion conveyed through narrative mechanisms such as conceptual incongruity or symbolic storytelling.
Recent work has begun documenting metric--human disagreement at scale: TIEdit~\cite{tiedit2026} finds limited correlation between automatic metrics and 307K human ratings for general image editing, and FED-Bench~\cite{fedbench2026} reveals ``lazy editing'' biases in facial expression metrics.
However, no prior work has stratified this disagreement \emph{by target emotion} within the emotion domain.
Our experiments fill this gap, revealing that classifier-based metrics systematically undervalue non-stereotypical, context-grounded edits that humans strongly prefer, with the blindness varying nearly an order of magnitude across target emotions.

\begin{figure*}[t]
    \centering
    \includegraphics[width=\linewidth]{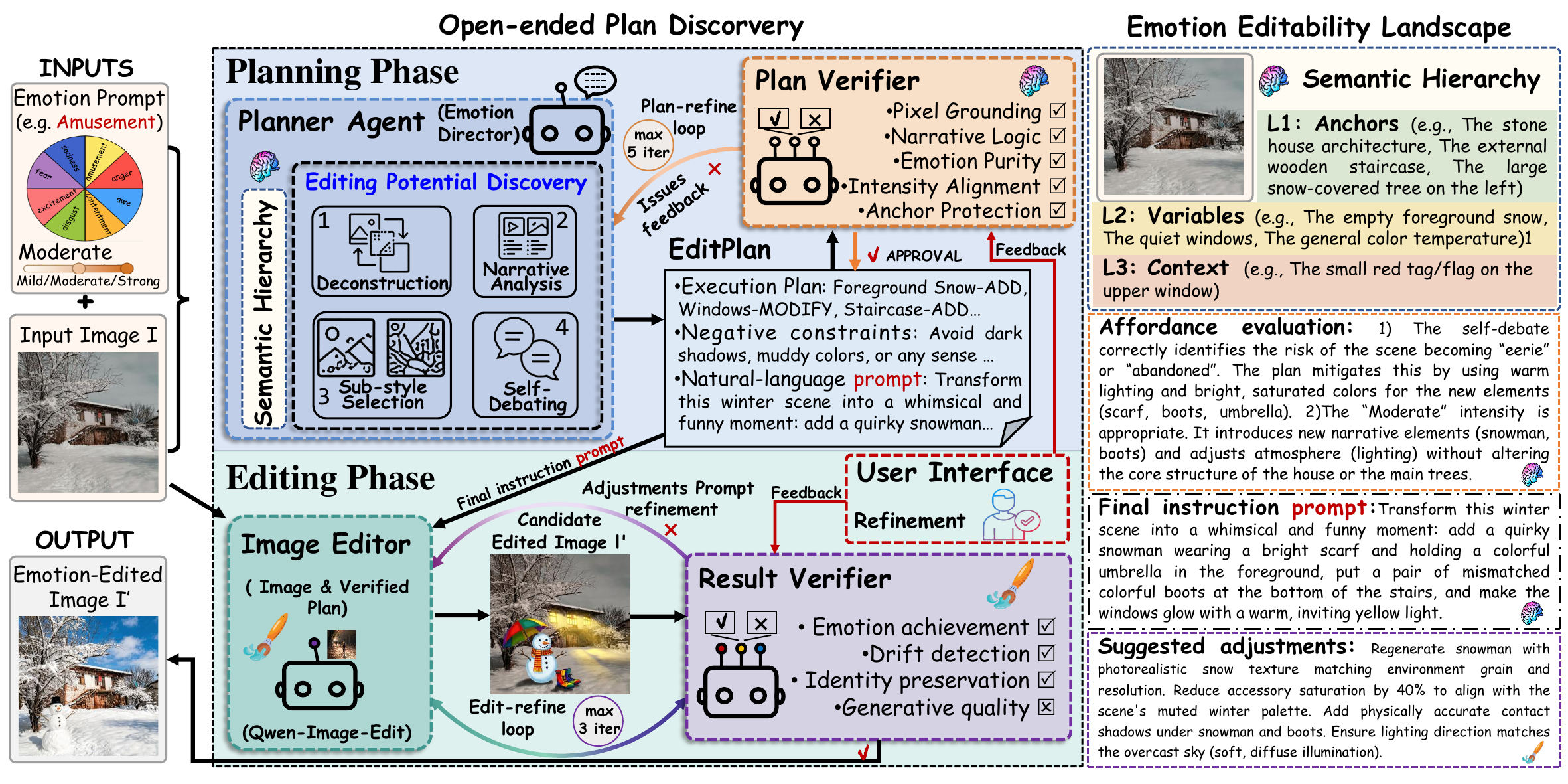}
    \caption{Overview of \method{}. Given an input image and a target emotion, the Planner discovers the editable space through emotion-conditioned reasoning and generates a structured editing plan. The Plan Verifier audits feasibility. The Image Editor executes the plan, and the Result Verifier evaluates emotional alignment. Both loops iterate until convergence.}
    \label{fig:architecture}
\end{figure*}

The question of \emph{per-image emotion editability}---whether every image is equally amenable to every target emotion---has received scattered attention.
EmoScene~\cite{emoscene2026} analyzes static correlations between affect, scene type, and perceptual attributes across 1.2M images, establishing that emotional associations vary by scene.
Theoretical and predictive editability work (Editing on the Generative Manifold~\cite{editmanifold2026}, ELECT~\cite{elect2025}, DreamSteerer~\cite{dreamsteerer2024}) treats editability as a model--image joint property at the timestep, manifold, or instance level.
These works collectively suggest that editability is image-dependent, but none directly quantifies how emotional editability varies across target emotions for a given source image.
Our analysis provides a first empirical proxy for this question by measuring how \method's relative human-preference advantage over baselines varies across image content and target emotion.


\section{Method}
\label{sec:method}

We present \method{}, an agent-based framework that discovers and exploits the full \emph{editable space} of each image through emotion-conditioned reasoning about its visual elements, letting a general-purpose vision-language model (VLM) generate open-ended editing strategies within that space.
The framework operates through two iterative loops orchestrated by four specialized agents: a \textbf{Planner}, a \textbf{Plan Verifier}, an \textbf{Image Editor}, and a \textbf{Result Verifier} (Fig.~{2}).

\subsection{Problem Formulation: The Editable Space}
\label{sec:editable_space}

Given an input image $I$ and a target emotion $e$ (e.g., \emph{amusement}), we define the \emph{editable space} $\mathcal{E}(I, e)$ as the set of all editing operations that satisfy two constraints:

\noindent\textbf{Consistency.}
The edited image $I'$ must remain recognizable as originating from $I$.
Crucially, consistency is not a single binary property but operates across multiple semantic levels: pixel-level (color fidelity), structural (spatial layout), identity (subject appearance), and narrative (thematic coherence).
Different levels of consistency preservation yield fundamentally different editing strategies.
For instance, preserving only narrative consistency---the viewer understands $I'$ as a variation of $I$'s story---permits far more expressive edits than requiring pixel-level fidelity.
This creates a \emph{consistency--expressiveness tradeoff}: relaxing consistency constraints expands the space of viable edits, while tightening them restricts the editor to surface-level adjustments.

\noindent\textbf{Plausibility.}
The edit must genuinely convey the target emotion $e$.
A technically flawless edit that evokes the wrong emotion---or whose intent is too absurd to interpret---fails this constraint regardless of its visual quality.

Formally, $\mathcal{E}(I, e) = \{ \delta \mid \mathcal{C}(I, I+\delta) \geq \tau_c \;\wedge\; \mathcal{P}(I+\delta, e) \geq \tau_p \}$, where $\mathcal{C}$ measures multi-level consistency and $\mathcal{P}$ measures emotional plausibility, with thresholds $\tau_c$ and $\tau_p$ that vary with the desired editing intensity.

\noindent\textbf{Contrast with prior work.}
Existing methods implicitly define $\mathcal{E}$ as a small, fixed set of editing categories predetermined at design time (\S{2}).
Our key insight is that $\mathcal{E}(I, e)$ is \emph{image-specific} and far larger than any fixed taxonomy can capture: making Vermeer's \emph{Girl with a Pearl Earring} more amusing might modify her expression (within traditional factor trees) or replace her with a cat wearing the same earring---a semantic-level conceptual substitution that no factor tree enumerates, yet humans immediately recognize as both a valid edit and genuinely funny.
We let the VLM discover $\mathcal{E}$ through emotion-conditioned reasoning about the image's narrative potential rather than prescribing it.

\subsection{Semantic Hierarchy}
\label{sec:hierarchy}

To operationalize the abstract $\mathcal{C}$/$\mathcal{P}$ constraints, we discretise consistency $\mathcal{C}$ into three semantic layers (Anchors, Variables, Context) with the threshold $\tau_c$ mapped to the editing-intensity setting (Mild / Moderate / Strong); plausibility $\mathcal{P}$ is enforced separately by the Result Verifier (Sec.~{3.5}).
The hierarchy thus defines \emph{editing boundaries}---what must not be changed at each intensity---rather than prescribing what should be changed:

\noindent\textbf{Layer~1: Anchors.}
Core identity elements that define ``what this image is'': face identity, specific landmarks, iconic objects.
These are \emph{never modified} unless the editing intensity is \textit{Strong}, ensuring identity preservation by default.

\noindent\textbf{Layer~2: Variables.}
Primary emotional vehicles: expressions, gaze, posture, lighting, and character interactions.
These are the default targets for emotional modulation.

\noindent\textbf{Layer~3: Context.}
The environmental ``stage'': background, weather, atmospheric conditions, and symbolic objects (e.g., an empty chair, a wilting flower).
These elements can be freely modified, added, or removed.

This hierarchy is fundamentally different from the factor trees used in prior work.
Factor trees enumerate \emph{what to edit}; our hierarchy defines \emph{what not to edit}.
Within the unconstrained space, the Planner freely discovers any strategy that satisfies consistency and plausibility.
The hierarchy also implements the consistency--expressiveness tradeoff through intensity levels:
\textit{Mild} restricts edits to Layer~2 micro-adjustments;
\textit{Moderate} permits contextual shifts across Layers~2--3;
\textit{Strong} unlocks Layer~1, enabling full narrative transformations.
Figure~{3} illustrates this ladder on two contrasting sources and surfaces a property that emerges naturally from the affordance-conditioned design: under the same target emotion, the sub-style chosen at each intensity level is itself a function of what the source image affords.


\begin{figure}[t]
    \centering
    \footnotesize
    \setlength{\tabcolsep}{2pt}
    \renewcommand{\arraystretch}{1.05}
    \begin{tabular}{@{}cccc@{}}
        \textbf{(a) Source} & \textbf{(b) Mild} & \textbf{(c) Moderate} & \textbf{(d) Strong} \\
        \includegraphics[width=0.245\linewidth]{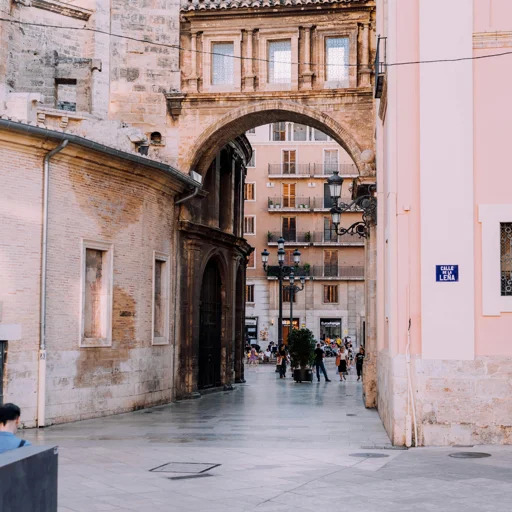} &
        \includegraphics[width=0.245\linewidth]{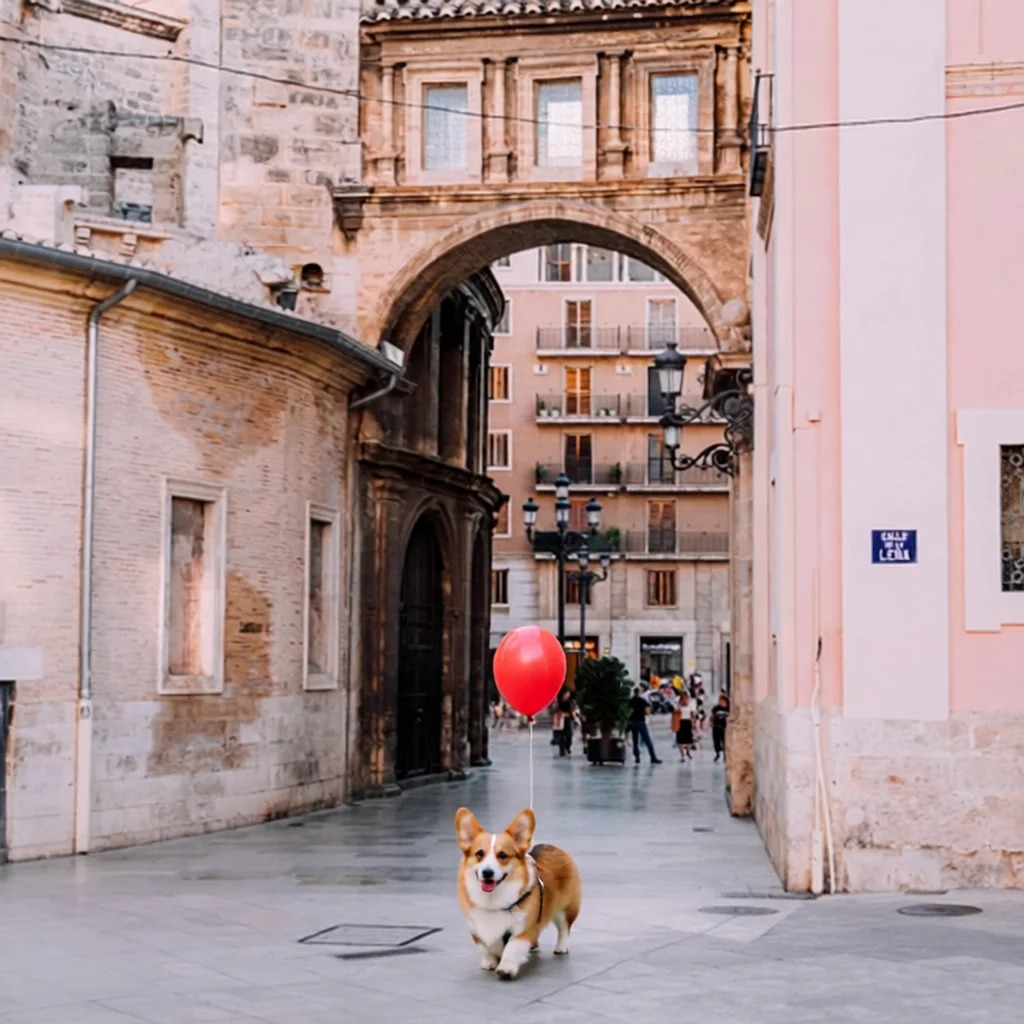} &
        \includegraphics[width=0.245\linewidth]{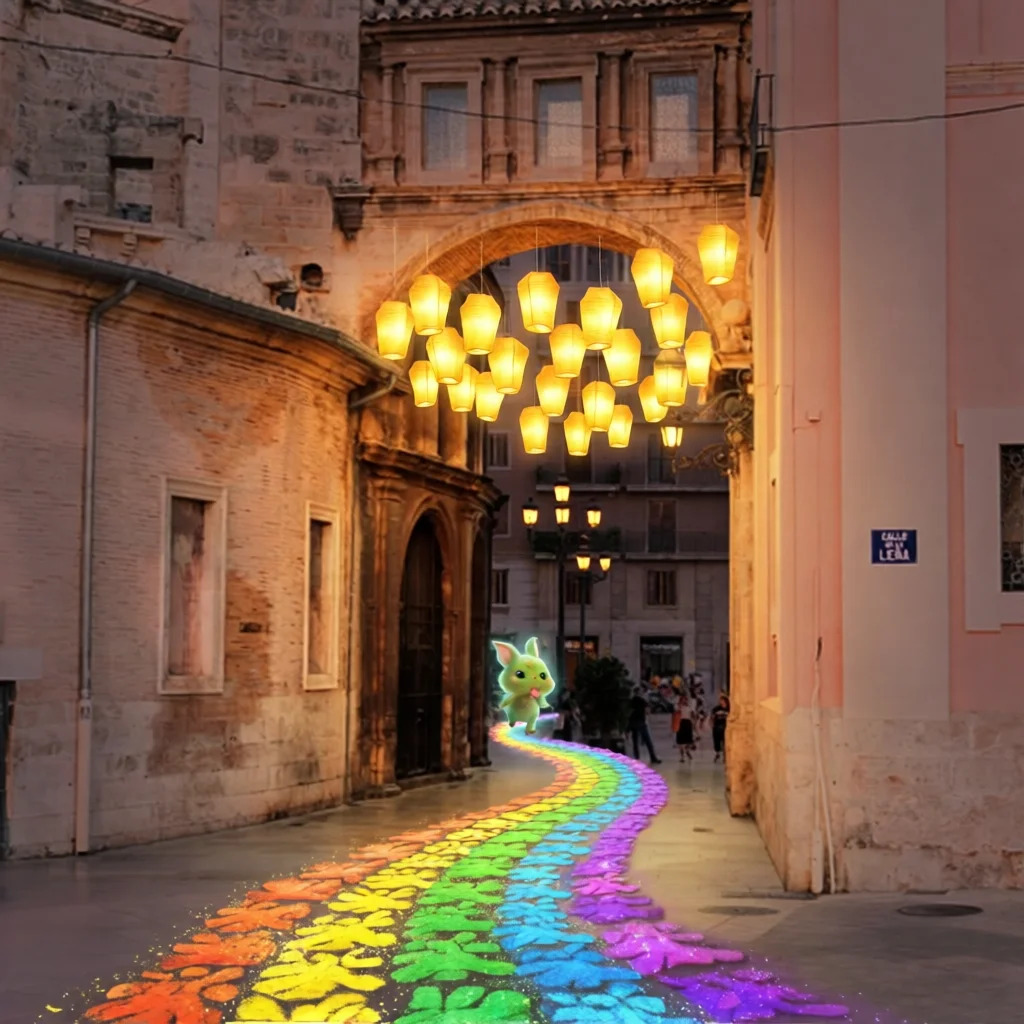} &
        \includegraphics[width=0.245\linewidth]{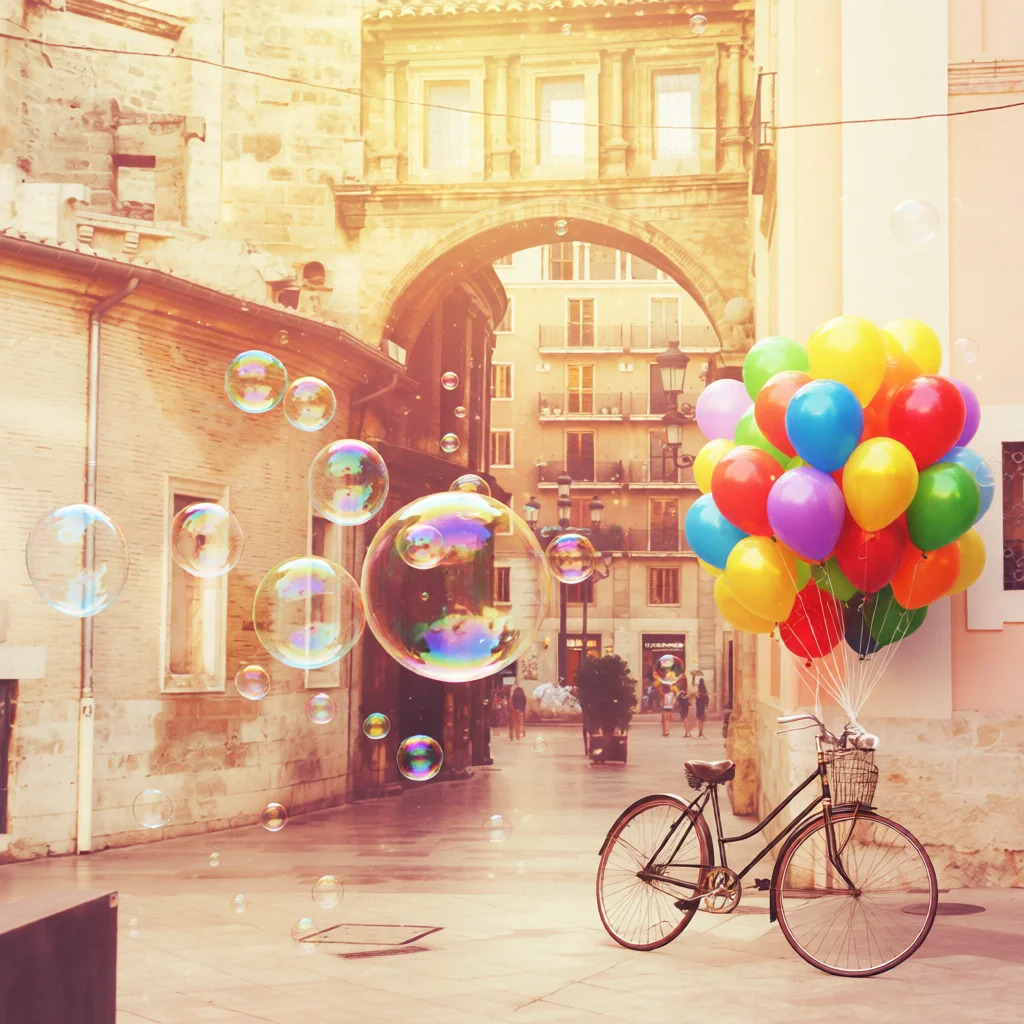} \\
         & \emph{Whimsical} & \emph{Whimsical} & \emph{Whimsical} \\[1pt]
        \includegraphics[width=0.245\linewidth]{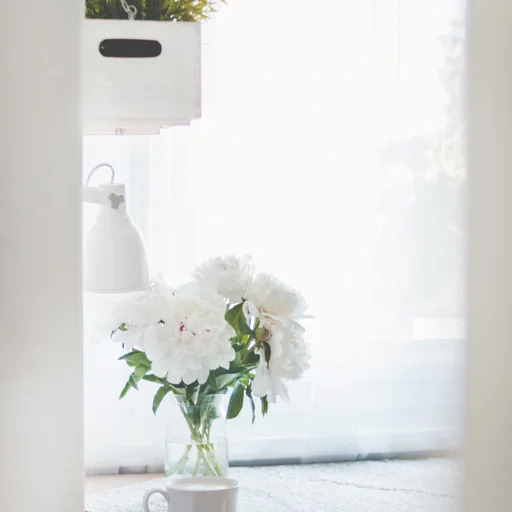} &
        \includegraphics[width=0.245\linewidth]{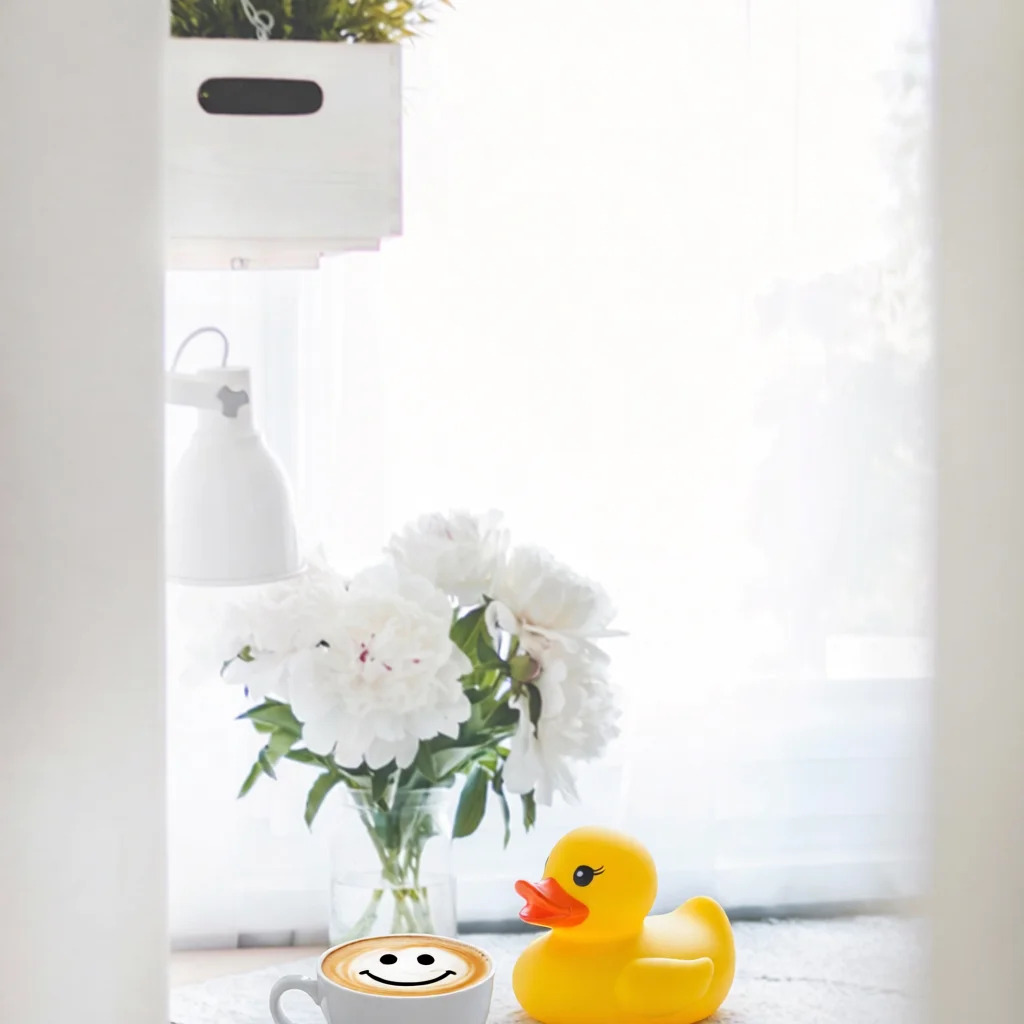} &
        \includegraphics[width=0.245\linewidth]{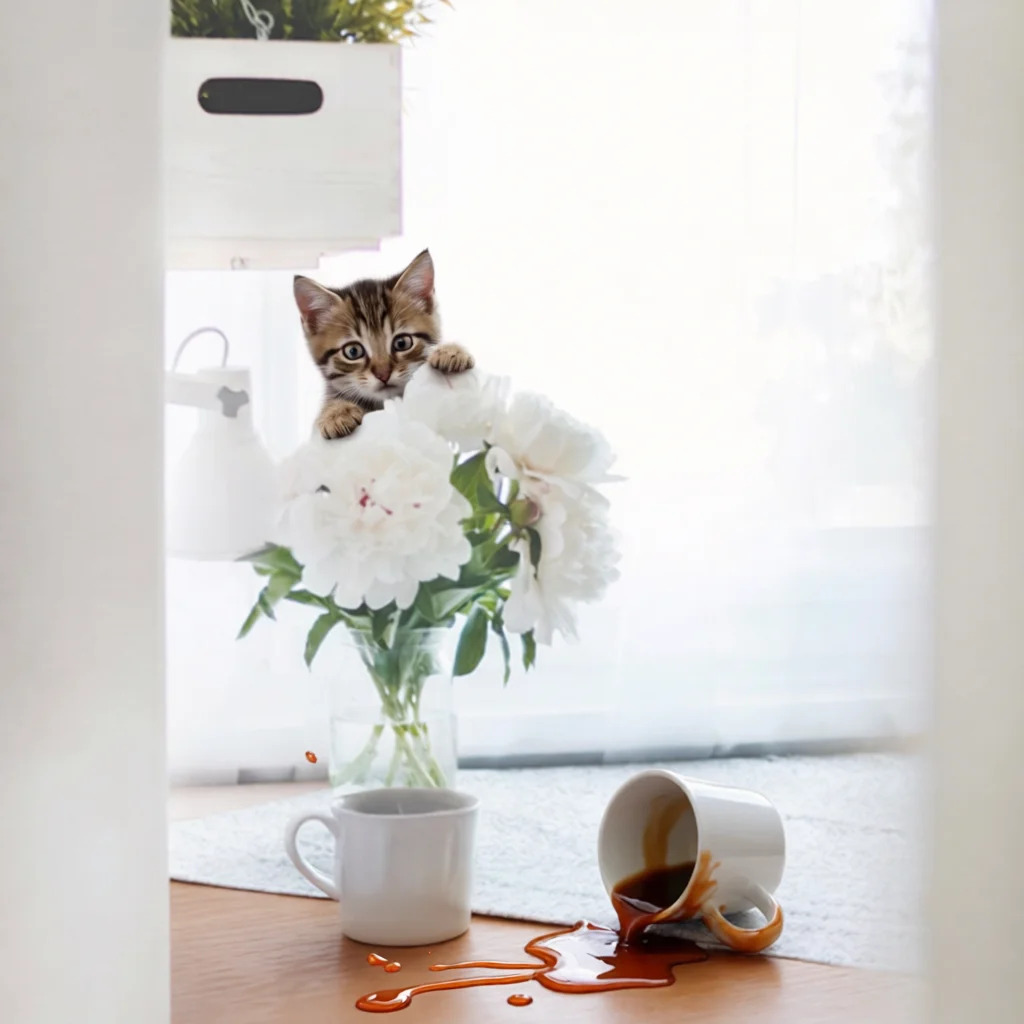} &
        \includegraphics[width=0.245\linewidth]{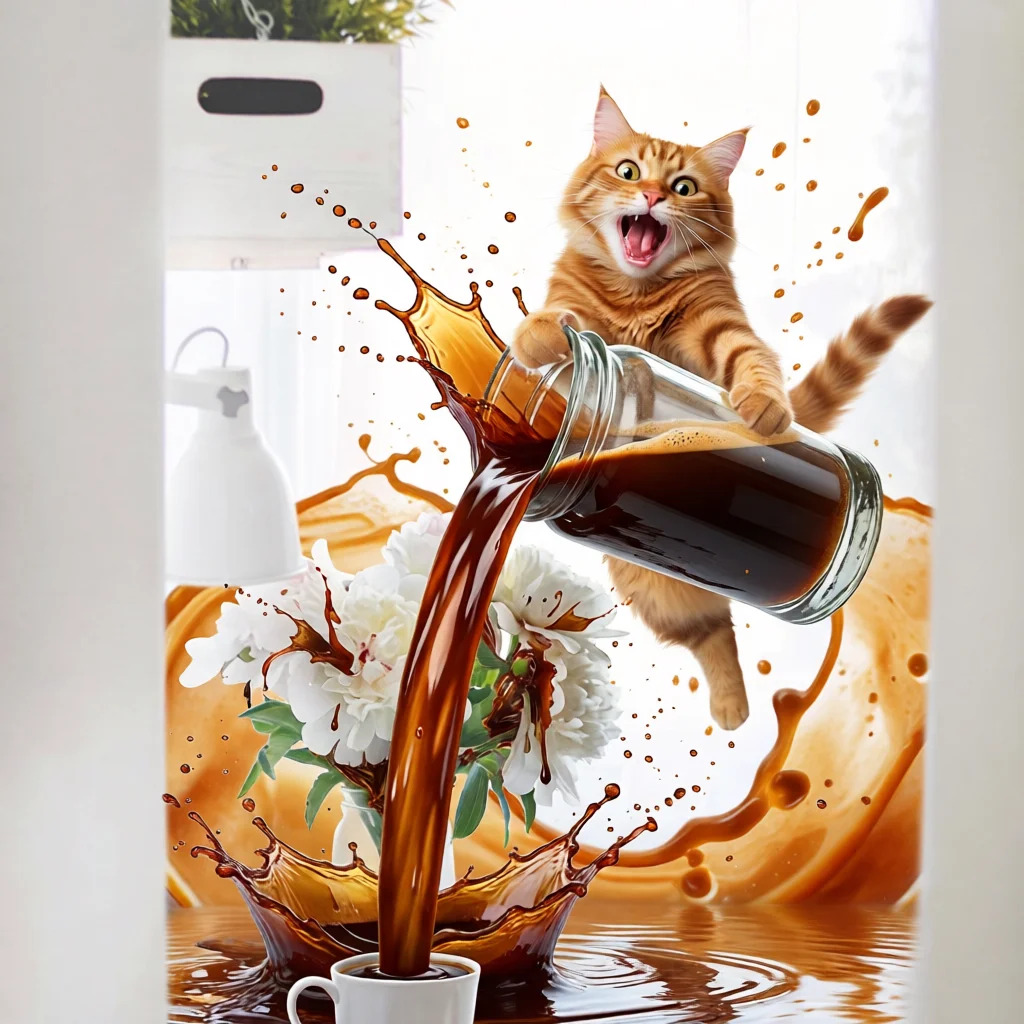} \\
         & \emph{Whimsical} & \emph{Slapstick} & \emph{Slapstick} \\
    \end{tabular}

    \caption{The semantic hierarchy defines \emph{editing boundaries}, not strategies; higher intensity unlocks deeper layers (Mild = Layer~2 only; Moderate = L2--3; Strong = full L1--3).
    Both rows target \emph{amusement}.
    Row~1 (street scene, narrative-heavy affordance) stays \emph{Whimsical} throughout: a balloon-carrying corgi (Mild) $\to$ hanging lanterns + chalk footprints + a small creature (Moderate) $\to$ warm storybook palette + giant bubbles + balloon-laden bicycle (Strong).
    Row~2 (still life, atmospheric-heavy) escalates from \emph{Whimsical} to \emph{Slapstick} as Layer~1 unlocks: a tiny duck and latte-art face (Mild) $\to$ tipped mug + guilty kitten (Moderate) $\to$ clumsy cat knocks vase and mug over in a bright splash (Strong).
    The sub-style adaptation across rows shows the Planner reading affordances rather than instantiating a fixed template.}
    \label{fig:hierarchy}
\end{figure}

\subsection{Editing Potential Discovery}
\label{sec:planner}

The Planner Agent receives the input image $I$, target emotion $e$, and intensity level, and outputs a structured \texttt{EditPlan} through a four-step reasoning protocol:

\noindent\textbf{Step~1: Deconstruction.}
The VLM analyzes $I$ and classifies every visible element into Anchors, Variables, or Dispensables.
This step answers: \emph{``What can I work with?''}

\noindent\textbf{Step~2: Narrative Evaluation.}
The VLM assesses whether $I$ has strong \emph{narrative potential}---whether visual elements suggest stories that could serve as emotional vehicles.
An empty park bench affords narratives of waiting or loneliness; a set dining table affords an interrupted meal.
Based on this assessment, the Planner selects one of three approaches:
\emph{Atmospheric} (modifying color, lighting, texture),
\emph{Narrative} (introducing or modifying story elements), or
\emph{Hybrid}.

\noindent\textbf{Step~3: Sub-style Selection.}
Each primary emotion is further differentiated into three sub-styles capturing distinct emotional nuances.
For example, \emph{amusement} branches into \emph{Slapstick} (goofy humor), \emph{Satirical} (ironic wit), and \emph{Whimsical} (playful magic).
The Planner selects the sub-style that best matches the image's narrative potential, yielding $8 \times 3 = 24$ fine-grained emotional targets across the eight Mikels categories~\cite{mikels2005emotional}.

\noindent\textbf{Step~4: Self-Debate.}
The Planner evaluates \emph{emotional drift risks}---the tendency for edits targeting one emotion to inadvertently evoke a neighboring one (e.g., high contrast intended for \emph{sadness} drifting into \emph{anger}).
This step produces negative constraints to prevent drift.

The output \texttt{EditPlan} contains: an execution plan specifying per-element actions (\texttt{MODIFY}, \texttt{ADD}, \texttt{DELETE}, or \texttt{KEEP}) with layer assignments; negative constraints; and a natural-language \texttt{final\_instruction\_prompt} written as a director would instruct an artist.

\subsection{Plan Verification}
\label{sec:plan_verifier}

Before execution, the Plan Verifier audits the \texttt{EditPlan} across five dimensions:
(1)~\emph{pixel grounding}---do elements marked for modification actually exist in $I$? Plans referencing non-existent elements are rejected as hallucinations;
(2)~\emph{narrative logic}---does the proposed story follow naturally from the scene's visual elements?
(3)~\emph{emotion purity}---are drift risks adequately addressed?
(4)~\emph{intensity alignment}---is edit scope appropriate for the requested level?
(5)~\emph{anchor protection}---are Layer~1 elements properly preserved?
Rejected plans return to the Planner with actionable feedback, forming the \emph{Plan-Refine loop} ($\leq K_p{=}5$ iterations).

\subsection{Image Editing and Result Verification}
\label{sec:editing}

The Image Editor receives the verified plan's \texttt{final\_instruction\_prompt} and applies it to $I$ using Qwen-Image-Edit~\cite{qwen_image_edit} as the execution engine.
Operating on natural-language instructions enables the open-ended strategies from our Planner without requiring specialized editing modules for each strategy type.

The Result Verifier then evaluates the edited image $I'$ by comparing it with both $I$ and the plan, assessing:
(1)~\emph{emotion achievement}---does $I'$ evoke the target emotion and sub-style?
(2)~\emph{drift detection}---has the image slipped into a neighboring emotion?
(3)~\emph{identity preservation}---are anchors intact?
(4)~\emph{generative quality}---are there artifacts?
Based on the assessment, the verifier issues one of three decisions:
\textsc{Approve},
\textsc{Retry\_Pixels} (re-edit with adjusted prompts), or
\textsc{Replan\_Strategy} (restart from the Planner, e.g., pivoting from Narrative to Atmospheric when the story fails to materialize).
This forms the \emph{Edit-Refine loop} ($\leq K_e{=}3$ iterations).

\subsection{Dual-Loop Architecture}
\label{sec:loops}

The complete pipeline chains the Plan-Refine and Edit-Refine loops described above.
The separation of planning from execution is critical: unlike single-pass methods that conflate strategy selection with image generation, our design ensures that editing strategy is grounded in actual image content before any pixels are modified, and that execution failures are distinguished from strategic failures for targeted recovery.


\section{Experiments}
\label{sec:experiments}

We organise our empirical study around six testable hypotheses (Tab.~{1}), each evaluated in a dedicated subsection; Sec.~{4.7} provides cross-source qualitative results that illustrate H1--H4 jointly.

\begin{table}[t]
\centering
\footnotesize
\setlength{\tabcolsep}{4pt}
\renewcommand{\arraystretch}{1.05}
\begin{tabular}{@{}llp{0.72\linewidth}@{}}
\toprule
\textbf{ID} & \textbf{\S} & \textbf{Hypothesis} \\
\midrule
H1 & {4.2}  & Open-ended planning produces edits that humans perceive as higher quality than prior AIM baselines. \\
H2 & {4.3}   & \method{} adapts its editing strategy to the target emotion rather than applying a uniform template. \\
H3 & {4.4}     & Classifier-based automated metrics fail systematically on non-stereotypical context-grounded edits, in an emotion-conditional manner. \\
H4 & {4.5} & \method's relative preference advantage varies systematically across image content and target emotion. \\
H5 & {4.6}    & Each component of the multi-agent pipeline contributes to overall performance. \\
H6 & {4.8} & Non-expert users can converge on personalised emotional edits within a small number of plan/edit iterations using \method's interactive workflow. \\
\bottomrule
\end{tabular}
\caption{The six hypotheses tested in this section.}
\label{tab:hypotheses}
\end{table}

\subsection{Experimental Setup}
\label{sec:exp_setup}

\noindent\textbf{Datasets and emotion taxonomy.}
For quantitative benchmarking we use two image sources spanning contrasting photographic regimes:
the EmoEdit benchmark~\cite{yang2024emoedit} (curated stock-style photographs, $\nFullEmoeditImg{}$ images),
and the EmoEditor benchmark~\cite{ling2024emoeditor} (in-the-wild photographic snapshots, $\nFullEmoeditorImg{}$ images).
We additionally use a $\nFullArtemisImg{}$-image painting subset drawn from ArtEmis~\cite{achlioptas2021artemis} as a \emph{qualitative} cross-source check (Fig.~{8})---it is an order of magnitude smaller than the two main benchmarks and we therefore do not include it in aggregate quantitative comparisons.
Each image is edited toward all eight emotion categories from the Mikels model~\cite{mikels2005emotional}: \emph{amusement, anger, awe, contentment, disgust, excitement, fear,} and \emph{sadness}, which combine the categorical view of basic emotions~\cite{ekman1992argument} with valence/arousal coverage in the spirit of Russell's circumplex~\cite{russell1980circumplex}.

\noindent\textbf{Evaluation splits.}
Different evaluations require different scales (Table~{2}), and we keep these splits clearly separated throughout the paper: automated metrics use the full quantitative benchmark ($\nFullBenchPairs{}$ outputs per method); human pairwise preference uses a stratified survey subset ($\nHumanImages{}$ images, $\nHumanQuestions{}$ pairwise questions); the metric--human correlation analysis joins these two at the (image, target emotion, baseline) level ($\nLinkedPairs{}$ paired records); ablation re-uses the EmoEdit benchmark; and qualitative figures additionally draw on a $\nFullArtemisImg{}$-image ArtEmis painting subset for cross-regime illustration only.
Throughout the paper, every reported $N$ corresponds to one of these splits and is named accordingly.

\begin{table}[t]
\centering
\caption{Evaluation protocol. Each row is a distinct split with its own scale and unit; the same number in two different rows refers to two different things. ArtEmis is qualitative-only; quantitative comparisons (Tab.~{7}) use only EmoEdit and EmoEditor due to ArtEmis's order-of-magnitude smaller scale.}
\label{tab:eval_protocol}
\resizebox{\linewidth}{!}{%
\begin{tabular}{lll}
\toprule
\textbf{Evaluation} & \textbf{Source images} & \textbf{Scale (per method)} \\
\midrule
Automated metrics & EmoEdit ($\nFullEmoeditImg{}$) + EmoEditor ($\nFullEmoeditorImg{}$) & $\nFullBenchPairs{}$ image--emotion outputs \\
Human pairwise preference & $\nHumanImages{}$ stratified-sampled subset & $\nHumanQuestions{}$ pairwise questions, $\nAnswers{}$ valid responses (target $\nDesignedResponses{}$)\\
Metric--human correlation & human-evaluated subset (joined with metrics) & $\nLinkedPairs{}$ paired records \\
Ablation & EmoEdit benchmark & $\nFullEmoeditPairs{}$ outputs $\times$ 5 configurations \\
Qualitative cross-regime check & ArtEmis paintings subset ($\nFullArtemisImg{}$) & illustrative only, no aggregates reported \\
\bottomrule
\end{tabular}}
\end{table}

\noindent\textbf{Baselines.}
We compare against two state-of-the-art methods:
\textbf{EmoEdit}~\cite{yang2024emoedit} (CVPR 2024), which constructs a hierarchical emotion factor tree and edits each factor independently;
and \textbf{EmoEditor}~\cite{ling2024emoeditor} (ICCV 2025), which uses a dual-branch architecture integrating global context with local semantic regions.

\noindent\textbf{Baseline scope and editor-confound control.}
The two quantitative baselines are released as complete diffusion-based AIM systems with their own trained editors and we evaluate them using their official code; \method{} uses Qwen-Image-Edit-2511~\cite{qwen_image_edit}.
A fair head-to-head against agentic AIM systems (EmoAgent~\cite{emoagent2025}, EmoFeedback$^2$~\cite{emofeedback2025}, Emotion-Director~\cite{emotiondirector2025}) at scale would also require unifying the generative backbone, which we leave to future work; instead we test the planning architecture's contribution \emph{within} our pipeline through the ``Editor only'' ablation (A5; Sec.~{4.6}), which runs raw Qwen-Image-Edit on the bare prompt ``Make this image evoke [emotion]'' with no planning, no verifier loops, and no agent at all.
Full \method{} substantially outperforms this same-editor baseline, evidence that the gains within our pipeline are not explained by the Qwen editor alone.
A qualitative system-level comparison against EmoAgent on four representative cases shows the same planning difference at the per-image level: generic agentic AIM tends to apply broad affective transformations, whereas \method{} grounds its edits in image-specific affordances.

\noindent\textbf{Implementation.}
All four agents (Planner, Plan Verifier, Image Editor controller, Result Verifier) use Gemini-3-Flash as the VLM backbone, accessed via the OpenRouter API; the image editor is Qwen-Image-Edit-2511~\cite{qwen_image_edit} in FP8-quantised form (using the publicly released checkpoint of \cite{qwen_image_edit}), invoked with $40$ inference steps and classifier-free guidance scale~$4.0$.
Default iteration limits are $K_p = 5$ (plan-refine) and $K_e = 3$ (edit-refine), with seed unfixed (system seed) and editing intensity \emph{Moderate} unless otherwise stated.
Inference runs on a single NVIDIA RTX-5090 GPU.
For reproducibility, we use Qwen-Image-Edit as the image editor and keep the Planner, Plan Verifier, and Result Verifier prompts fixed across all benchmark runs; dataset licences follow the original EmoEdit, EmoEditor, and ArtEmis releases.

\noindent\textbf{User Study Design.}
The human pairwise preference study uses a $2$-alternative forced-choice (2AFC) design.
Each of the $\nHumanTasks{}$ image--emotion tasks (Sec.~{4.1}, Table~{2}) is evaluated against the two baselines, producing $\nHumanQuestions{}$ pairwise questions.
Each question is assigned to at least three independent raters (a $\nDesignedResponses{}$-response design target).
Each question contains four sub-questions:
\textbf{Q1}~(Emotion Accuracy),
\textbf{Q2}~(Visual Quality, with a ``both unacceptable'' option),
\textbf{Q3}~(Identity Balance, with failure tagging), and
\textbf{Q4}~(Dynamic Attribution: multi-select from \emph{Global Atmosphere}, \emph{Narrative Element}, \emph{Subject Behavior}, \emph{Other}).
We collected $\nRawAnswers{}$ raw responses from $\nRawParticipants{}$ participants---above the $\nDesignedResponses{}$-response design target ($\nHumanQuestions{}$ questions $\times$ $3$ raters)---and applied two quality filters: sessions with any failed attention check are excluded, and sessions with per-question speed below 20 seconds are excluded.
After filtering, $\nAnswers{}$ valid responses across $\nSessions{}$ valid sessions from $\nUniqueParticipants{}$ unique participants are retained for analysis (a $\pctFilteredOut{}$ rejection rate); the same participant may complete more than one session, and statistical inference (Sec.~{4.2}) accounts for this nesting via clustered bootstrapping over source images.
The questionnaire presents source, baseline, and \method{} outputs in randomized order and asks raters to judge emotional preference, image quality, identity preservation, and the visual source of the preferred emotion; attention checks remove sessions with inconsistent or incomplete answers.

\noindent\textbf{Automated Metrics.}
We report Emo-A (emotion classification accuracy: $1$ if a pretrained 8-way Mikels-emotion classifier predicts the target emotion for the edited image, else $0$), Emo-S (the classifier's softmax confidence on the target class), CLIP-I (image--image cosine similarity in CLIP-ViT-L/14 feature space~\cite{radford2021clip}), and LPIPS~\cite{zhang2018unreasonable} (perceptual distance from the original).
The emotion classifier is the pretrained model released with EmoSet~\cite{emoset2023}, which is the standard reference classifier used by both EmoEdit and EmoEditor.

\subsection{Human Preference Evaluation}
\label{sec:human_pref}

Table~{3} reports the human preference rate across three evaluation dimensions.
Our method achieves \qOneOverall{} overall preference on Q1, averaged across the two baseline comparisons.
All cells are significant at $p < .001$ under a two-sided binomial test; the primary Q1 result also survives a clustered bootstrap that resamples source images to account for the nested vote structure ($95\%$ CIs $\qOneVsEmoeditCI$ vs EmoEdit and $\qOneVsEmoeditorCI$ vs EmoEditor, both far above the $50\%$ no-preference threshold).

\begin{table}[t]
\centering
\caption{Human preference rates (\%). Values above 50\% indicate preference for our method; all cells $p < .001$ (two-sided binomial test). For the Q1 row only, clustered-bootstrap $95\%$ CIs (resampling source images, $B=2{,}000$) are $\qOneVsEmoeditCI$ and $\qOneVsEmoeditorCI$ for EmoEdit and EmoEditor respectively, confirming significance under the nested-vote structure.}
\label{tab:human_pref}
\begin{tabular}{lcc}
\toprule
\textbf{Dimension} & \textbf{vs.\ EmoEdit} & \textbf{vs.\ EmoEditor} \\
\midrule
  Q1: Emotion Accuracy & \textbf{83.0}\% $p<.001$ & \textbf{93.1}\% $p<.001$ \\
  Q2: Visual Quality & \textbf{74.6}\% $p<.001$ & \textbf{87.1}\% $p<.001$ \\
  Q3: Identity Balance & \textbf{70.3}\% $p<.001$ & \textbf{86.9}\% $p<.001$ \\
\bottomrule
\end{tabular}
\end{table}

Table~{4} breaks down Q1 by emotion direction.
Our method achieves consistently strong preference across all eight emotions (ranging from \qOneAweVsEditor{} to \qOneExcitementVsEditor{} against EmoEditor), with the highest margins against EmoEditor on \emph{excitement} (\qOneExcitementVsEditor{}), \emph{contentment} (\qOneContentmentVsEditor{}), and \emph{anger} (\qOneAngerVsEditor{})---emotions that benefit most from our method's open-ended strategy discovery.

\begin{table}[t]
\centering
\caption{Human preference rates (\%) by emotion direction (Q1: Emotion Accuracy).}
\label{tab:pref_by_emotion}
\begin{tabular}{lcc}
\toprule
\textbf{Emotion} & \textbf{vs.\ EmoEdit} & \textbf{vs.\ EmoEditor} \\
\midrule
  Amusement & 79.0\% & 93.2\% \\
  Anger & 71.8\% & 95.3\% \\
  Awe & 81.3\% & 86.9\% \\
  Contentment & 83.9\% & 95.9\% \\
  Disgust & 80.9\% & 93.5\% \\
  Excitement & 91.0\% & 98.0\% \\
  Fear & 86.8\% & 91.7\% \\
  Sadness & 89.2\% & 90.1\% \\
\bottomrule
\end{tabular}
\end{table}

To verify the robustness of these preferences, we examine Q1--Q2--Q3 coherence (Table~{5}).
When restricting to votes where Q1, Q2, and Q3 all agree on the same method, our method achieves a \coherentOurs{} preference rate, confirming that our advantage holds across all evaluation dimensions simultaneously.

\begin{table}[t]
\centering
\caption{Q1--Q2--Q3 coherence analysis. ``Sweeps'' indicates all three dimensions favor the same method.}
\label{tab:coherence}
\begin{tabular}{lcc}
\toprule
\textbf{Pattern} & \textbf{Count} & \textbf{\%} \\
\midrule
  Ours sweeps Q1+Q2+Q3 & 2755 & \textbf{58.7}\% \\
  Baseline sweeps & 238 & 5.1\% \\
  Mixed & 1700 & 36.2\% \\
\midrule
  Coherent-only ours pref. & 2755/2993 & \textbf{92.0}\% \\
\bottomrule
\end{tabular}
\end{table}

\noindent\textbf{Robustness across participant demographics.}
We further verify that the preference for \method{} is not driven by any single demographic subgroup.
Stratifying the Q1 votes by \nDemoNDims{} self-reported participant attributes (age range, field of study, AI familiarity, visual-art experience, and self-rated emotion sensitivity), the lowest stratum-level preference rate is \nDemoMinPref{} (subgroup: \nDemoMinSubgroup{}, $N_{\text{resp}}=\nDemoMinN{}$) and the highest is \nDemoMaxPref{} (\nDemoMaxSubgroup{}); no demographic stratum prefers a baseline on average.
This robustness holds across the collected demographic strata, including age, field of study, AI familiarity, visual-art experience, and self-rated emotion sensitivity.

The combined preference, coherence, statistical, and demographic-robustness evidence collectively supports \textbf{H1}.

\subsection{Strategy Adaptivity and Open-Ended Edit Patterns}
\label{sec:diversity}

If \method's edits draw from a richer set of strategies than the fixed taxonomies of prior methods, this should manifest as \emph{target-emotion-conditional} variation in which kinds of edits humans perceive.
We test this using the Q4 (Dynamic Attribution) responses, which capture which kind of change participants felt drove each method's edit (Global Atmosphere, Narrative Element, Subject Behavior, Other).

\noindent\textbf{Aggregate distributions are similar; per-emotion distributions are not.}
Table~{6} compares the marginal attribution distribution between ours' and baselines' winning edits.
The two are visually similar (atmosphere-dominated, narrative-second), and we deliberately \emph{do not} read this aggregate as evidence of greater raw diversity for ours---marginal entropies of attribution categories are too coarse a measure to distinguish a fixed taxonomy from an open-ended one.
The informative quantity is the \emph{conditional} distribution: how does the Q4 mix shift with target emotion (Fig.~{4})?

\begin{table}[t]
\centering
\caption{Q4 attribution distribution (\%) when each method wins Q1. $\Delta$ = ours $-$ baseline. Marginals are similar; the per-emotion conditional distribution (Fig.~{4}) is where the adaptive behaviour shows.}
\label{tab:attribution}
\begin{tabular}{lccc}
\toprule
\textbf{Attribution} & \textbf{Ours wins} & \textbf{BL wins} & \textbf{$\Delta$} \\
\midrule
  Global Atmosphere & 46.0\% & 46.5\% & -0.5\% \\
  Narrative Element & 32.1\% & 32.8\% & -0.7\% \\
  Subject Behavior & 20.5\% & 15.5\% & +5.0\% \\
  Other & 1.5\% & 5.2\% & -3.8\% \\
\midrule
  Shannon $H$ & 1.600 & 1.681 & \\
\bottomrule
\end{tabular}
\end{table}

\begin{figure}[t]
\centering
\includegraphics[width=\linewidth]{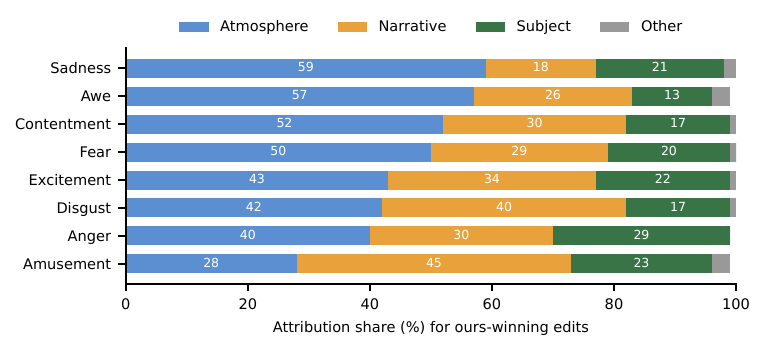}
\caption{Q4 attribution distribution (\%) for \method's winning edits, by target emotion (rows ordered by Atmosphere share so the shift is visible): \emph{amusement}/\emph{disgust} rely on narrative strategies, \emph{sadness}/\emph{awe}/\emph{contentment} are atmosphere-dominated, \emph{anger} is balanced.}
\label{fig:q4_by_emotion}
\end{figure}

\noindent\textbf{Reading.}
Three patterns are evident.
(i)~Narrative-element rate spikes for \emph{amusement}~($45\%$) and \emph{disgust}~($40\%$), the two emotions most amenable to surprise/incongruity-based interpretations.
(ii)~Global atmosphere dominates \emph{sadness}~($59\%$), \emph{awe}~($57\%$), and \emph{contentment}~($52\%$), where mood is conveyed primarily by lighting, palette, and setting.
(iii)~\emph{Anger} alone exhibits a balanced $40/30/30$ split across the three primary categories, consistent with the fact that anger admits multiple perceptually-distinct visual idioms (red light, sharp expression, violent context).
We interpret this as evidence that \method{} \emph{adapts} its strategy to the target emotion rather than applying a uniform pipeline---a property that requires the planner to reason about per-image affordances rather than instantiate a fixed template.

We do \emph{not} claim from this analysis alone that ours' raw \emph{strategy vocabulary} is larger than that of prior AIM baselines: the Q4 four-bucket coding is too coarse to make that comparison rigorously.
The qualitative cross-source figure (Fig.~{8}) and the per-emotion Case~A analysis in Sec.~{4.4} provide the complementary evidence that ours produces edits outside the visual vocabulary favored by classifier-based metrics, and qualitatively different from the stereotyped baseline outputs we observe.

We additionally collected per-response bounding-box annotations of the regions humans attributed each edit to (Q4.1) as raw data for future analysis (see \S{5}); these annotations are most meaningful at the per-image level rather than as a method-aggregated heatmap, and we leave a proper per-image trigger-region analysis to future work.

The emotion-conditional shift in attribution distribution supports \textbf{H2}: \method{} selects context-appropriate strategies as a function of target emotion rather than executing a fixed pipeline.

\subsection{Why Current Metrics Fail: Stereotype Matching vs. Semantic Understanding}
\label{sec:metrics}

Table~{7} shows that \method{} attains the best aggregate scores on Emo-A, Emo-S, and CLIP-I across both quantitative datasets; per-emotion analysis further shows that the agreement between automatic metrics and human preference is highly target-dependent.
However, aggregate scores can hide systematic patterns of metric--human disagreement.
Because our questionnaire collects per-(image, target emotion) human preferences and our objective evaluation provides per-(image, target emotion) metric scores, we can investigate this disagreement directly: at what granularity do automated metrics actually agree with human judgment?

\begin{table}[t]
\centering
\caption{Aggregate automated metrics on the two quantitative benchmarks ($\nFullBenchPairs{}$ image--emotion outputs per method, summed across EmoEdit and EmoEditor). \method{} achieves the best Emo-S, Emo-A, and CLIP-I on both datasets; LPIPS (lower is better) reflects the larger semantic edits required by our open-ended strategy discovery. Best per column in \textbf{bold}.}
\label{tab:auto_metrics}
\resizebox{\linewidth}{!}{
\begin{tabular}{lcccc|cccc}
\toprule
\textbf{Method} & \multicolumn{4}{c|}{\textbf{EmoEdit} (415 imgs $\times$ 8 emos)} & \multicolumn{4}{c}{\textbf{EmoEditor} (504 imgs $\times$ 8 emos)} \\
\cmidrule(lr){2-5} \cmidrule(lr){6-9}
 & LPIPS$\downarrow$ & CLIP-I$\uparrow$ & Emo-S$\uparrow$ & Emo-A$\uparrow$ & LPIPS$\downarrow$ & CLIP-I$\uparrow$ & Emo-S$\uparrow$ & Emo-A$\uparrow$ \\
\midrule
  EmoEdit~\cite{yang2024emoedit} & \textbf{0.287} & 0.814 & 0.317 & 0.47 & \textbf{0.267} & 0.828 & 0.227 & 0.37 \\
  EmoEditor~\cite{ling2024emoeditor} & 0.398 & 0.776 & 0.254 & 0.25 & 0.33 & 0.676 & 0.212 & 0.26 \\
  AIF~\cite{weng2023aif} & 0.51 & 0.624 & 0.026 & 0.16 & 0.465 & 0.813 & 0.018 & 0.14 \\
  CLVA~\cite{fu2021clva} & 0.41 & 0.773 & 0.049 & 0.18 & 0.404 & 0.717 & 0.046 & 0.17 \\
  SDEdit~\cite{meng2021sdedit} & 0.442 & 0.631 & 0.212 & 0.34 & 0.476 & 0.633 & 0.207 & 0.34 \\
  \textbf{\method{}} & 0.322 & \textbf{0.823} & \textbf{0.338} & \textbf{0.52} & 0.351 & \textbf{0.848} & \textbf{0.333} & \textbf{0.5} \\
\bottomrule
\end{tabular}
}
\end{table}

\noindent\textbf{Setup.}
We link each metric score to the corresponding human Q1 preference, yielding $N=\nLinkedPairs{}$ paired records of (image, target emotion, baseline) and compute the Pearson correlation between the metric difference (ours $-$ baseline) and ours' human win rate.

\noindent\textbf{Subset bias.}
The survey subset is non-uniformly sampled relative to the full benchmark: the EmoEdit baseline's Emo-A is $58.1\%$ on the survey subset versus $47.7\%$ on the full benchmark (a $+10$~pp shift toward visually canonical, classifier-friendly examples), while ours' Emo-A is flat ($51.9\%$ vs $51.3\%$).
This bias \emph{strengthens} what follows: where the metric most agrees with the baseline, humans still prefer ours by a large margin.

\noindent\textbf{Finding 1: aggregate correlation is weak.}
Table~{8} shows that even the best-aligned metric, Emo-S, achieves only $r=+0.22$ ($\sim 5\%$ explained variance), and CLIP-I shows essentially zero correlation.
These results indicate that current classifier-based metrics are insufficient as standalone per-image proxies for human judgment.

\begin{table}[t]
\centering
\caption{Correlation between objective metric differences (ours $-$ baseline) and human Q1 win rate, with bootstrap 95\% CI ($N=\nLinkedPairs{}$ paired records). $^{\ast}$~indicates the CI excludes zero.}
\label{tab:metric_correlation_overall}
\begin{tabular}{ccc}
\toprule
\textbf{Metric (ours $-$ baseline)} & \textbf{Pearson $r$} & \textbf{95\% CI} \\
\midrule
  Emo-A & $+0.192$$^{\ast}$ & $[+0.144, +0.243]$ \\
  Emo-S & $+0.221$$^{\ast}$ & $[+0.171, +0.271]$ \\
  CLIP-I & $-0.043$ & $[-0.090, +0.003]$ \\
  LPIPS & $+0.075$$^{\ast}$ & $[+0.029, +0.123]$ \\
\bottomrule
\end{tabular}
\end{table}

\noindent\textbf{Finding 2: per-emotion reliability is highly heterogeneous.}
Figure~{5} stratifies the correlation by target emotion. The pattern is dramatic: for \emph{anger} the metric strongly predicts human judgment (Emo-A $r=\bestMetricR{}$, tight CI); for \emph{awe} the metric shows \emph{no significant agreement} (Emo-A $r=\worstMetricR{}$, $95\%$ CI $[-0.23, +0.03]$).
The remaining six cluster between, all weakly positive.
Aggregating across emotions silently averages this 0.6-wide swing into a single misleading number.

\begin{figure}[t]
\centering
\includegraphics[width=\linewidth]{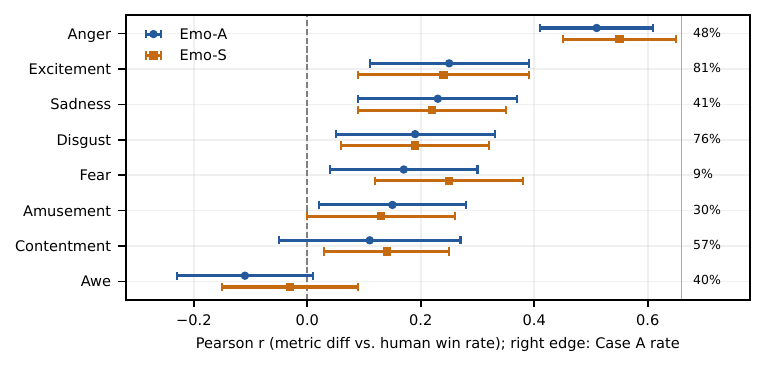}
\caption{Per-emotion metric--human correlation: Pearson $r$ of metric diff vs.\ human win rate (Emo-A, Emo-S) with $95\%$ bootstrap CIs; dashed line at $r=0$. Right column shows Case~A rate (\% of human-loved-ours edits Emo-A rejects). Emotions ordered by Emo-A reliability. Anger's CI sits above zero; awe's straddles zero; Case~A rate spans $9\times$ ($9\%$ fear $\to 81\%$ excitement). Pearson--Spearman shift $\leq \maxPearsonSpearmanDiff$.}
\label{fig:metric_per_emotion}
\end{figure}

\noindent\textbf{Finding 3: per-emotion metric blindness varies dramatically.}
We define \emph{Case~A} as a pair where the metric rejects ours (ours\_emo\_a$=0$, baseline\_emo\_a$=1$) but humans prefer ours (win rate $\geq 0.66$).
The per-emotion Case~A rate (right panel, Fig.~{5}) spans nearly an order of magnitude---from $9\%$ (fear) to $81\%$ (excitement) and $76\%$ (disgust)---a $9\times$ spread tracking the correlation pattern in Finding~2.
We also observe an absolute Case~A:Case~B asymmetry (\caseAcount{}:\caseBcount{}~events), but this largely reflects ours' overall preference advantage---under the null that metric outcome is independent of human winner, the marginal $87\%$ ours-preference rate alone predicts $\sim 6\times$ asymmetry without any metric bias.
The directly informative quantity is therefore the \emph{per-emotion} Case~A rate, not the aggregate ratio.

\noindent\textbf{Finding 4: the blindness is emotion-driven, not strategy-driven.}
A natural hypothesis is that the classifier is specifically blind to a particular kind of edit (e.g.\ narrative-level changes) and well-aligned on others (e.g.\ atmospheric tweaks).
We tested this by stratifying Case~A frequency by the dominant Q4 attribution category and found a tight cluster across Global~Atmosphere, Narrative~Element, and Subject~Behavior categories ($8.3$--$9.1\%$, $<1$~pp spread)---two orders of magnitude smaller than the per-emotion spread reported in Finding~3.
The metric blindness is therefore \emph{emotion-conditional}, not \emph{strategy-conditional}: it is the target emotion (and the breadth of valid visual idioms it admits) that determines whether the classifier will recognise ours' edit, not the kind of editing operation performed.

\noindent\textbf{Interpretation.}
The four findings are consistent with a single mechanism: current emotion classifiers behave as \emph{stereotype matchers}, not as semantic understanders. They match human perception well when training-distribution cues are concentrated and visually distinctive (fear: darkness; anger: red, sharp expression), and fail when valid perceptual cues are diverse, contextual, or conceptual (awe, excitement, disgust). Our open-ended strategy discovery produces many such non-stereotypical edits (Fig.~{6}), generating the systematic asymmetry observed.

\begin{figure}[t]
\centering

\providecommand{\oursbox}[1]{%
  {\setlength{\fboxrule}{1.5pt}\setlength{\fboxsep}{0pt}%
   \fcolorbox{oursaccent}{white}{#1}}%
}
\providecommand{\baselinebox}[1]{%
  {\setlength{\fboxrule}{0.6pt}\setlength{\fboxsep}{0pt}%
   \fcolorbox{black!55}{white}{#1}}%
}

\newcommand{\caseAcell}[6]{%
  \begin{tabular}{@{}c@{\hspace{2pt}}c@{\hspace{2pt}}c@{}}
    \multicolumn{3}{c}{\footnotesize\textbf{#1}} \\[1pt]
    \includegraphics[width=0.30\linewidth]{#2} &
    \baselinebox{\includegraphics[width=0.30\linewidth]{#3}} &
    \oursbox{\includegraphics[width=0.30\linewidth]{#4}} \\[1pt]
    {\scriptsize Source} &
    {\scriptsize\color{black!70}Baseline\,\cmark\;Emo-S\;$#5$} &
    {\scriptsize\color{oursaccent}Ours\,\xmark\;Emo-S\;$#6$} \\
  \end{tabular}%
}

{\renewcommand{\arraystretch}{0}\setlength{\tabcolsep}{0pt}%
\begin{tabular}{@{}c@{}}
  \caseAcell{Excitement}{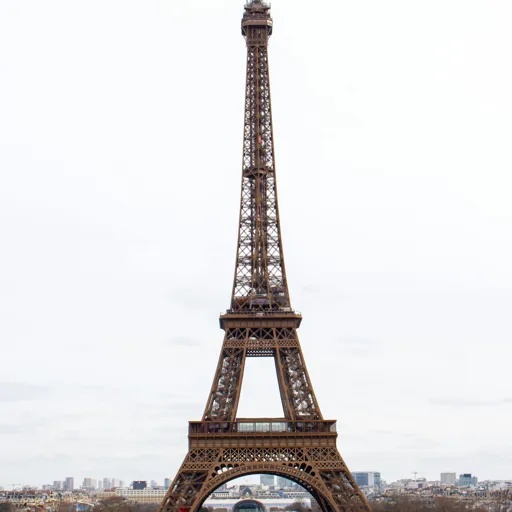}{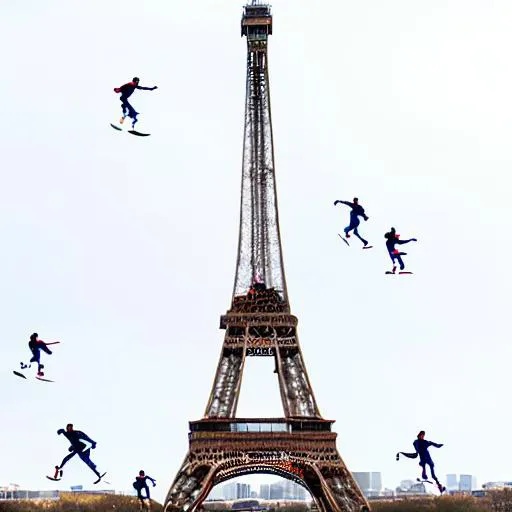}{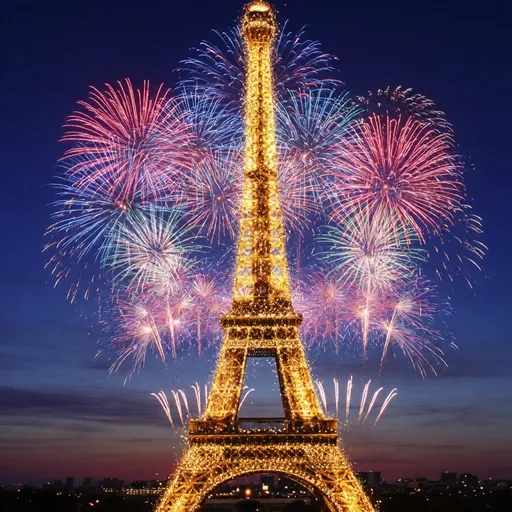}{+0.74}{+0.01} \\
\end{tabular}}

    \caption{\textbf{One Case~A example (Excitement).} Source, baseline's edit (Emo-A \cmark, classifier accepts), our edit (Emo-A \xmark, classifier rejects); human raters unanimously prefer ours (3/3). The baseline instantiates a stereotype template (pasted action-sport figures); ours produces a context-grounded interpretation (an Eiffel Tower fireworks night) that the classifier fails to read. This case is representative of the broader Case~A pattern: automatic classifiers often reward familiar affective stereotypes while humans prefer context-grounded edits.}
    \label{fig:metric_disagreement}
\end{figure}

\noindent\textbf{Implication for the field.}
Three implications follow:
(i)~per-emotion calibration is essential before automated metrics can substitute for human evaluation;
(ii)~the classification paradigm itself is limiting---a classifier trained on labelled stereotypes cannot, in principle, recognise edits that succeed through non-stereotypical means;
(iii)~a research direction opens for VLM-based evaluators that can reason about \emph{why} an image evokes an emotion, not just \emph{whether} its features match a stereotype.

The findings above jointly support \textbf{H3}.

\subsection{Relative Emotion Editability: Content--Emotion Preference Affinity}
\label{sec:editability}

We investigate a question that, to our knowledge, has not been directly addressed in prior work: \emph{is the relative human-preference advantage of our method uniform across image content and target emotion, or does it vary systematically?}
We group source images by visual content type (portrait, landscape, painting, architecture, urban, indoor, animal, object, action) and compute our Q1 preference rate for each (content type, target emotion) pair, yielding a $9 \times 8$ landscape across 72 cells.
We emphasise upfront that this measures \emph{relative} preference of ours over baselines, not the \emph{absolute} editing difficulty of each cell---a low cell value can reflect either ``hard for everyone'' or ``baseline happens to do well here''. We approximate absolute editability through this comparative landscape; an absolute Likert measure is left to future work.

Figure~{7} reveals a clear \emph{content--emotion preference affinity landscape}: \method's advantage varies systematically with both image content and target emotion.
\emph{Excitement} is the easiest target (95\% mean, $\geq 89\%$ on 8/9 content types); \emph{portraits} (91\% avg.) and \emph{urban} scenes (92\%) are uniformly editable.
In contrast, \emph{anger} shows the strongest content-dependence: nearly perfect on paintings (93\%) but degraded on landscapes (69\%) and architecture (73\%).
\emph{Indoor} scenes span $74\%$--$97\%$ across emotions; \emph{objects} excel for positive emotions but resist \emph{fear} (67\%, everyday items resist becoming menacing); \emph{action} scenes reach $100\%$ for both \emph{fear} and \emph{sadness}.
This within-category variance is the signature of the preference-affinity landscape: relative advantage is a joint function of content and emotion, not an intrinsic image property.

\begin{figure}[t]
    \centering
    \includegraphics[width=\linewidth]{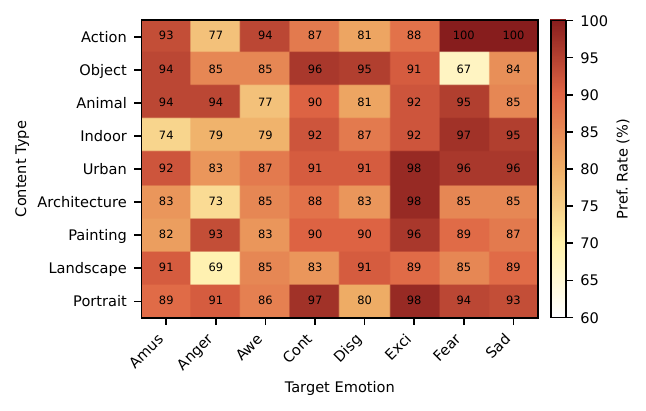}
    \caption{Content--emotion preference-affinity landscape ($9 \times 8$): \method's Q1 preference rate (\%) by image content type (row) $\times$ target emotion (column). \emph{Relative} measure of advantage over baselines, not absolute editability.}
    \label{fig:editability}
\end{figure}

The systematic non-uniformity of cell values across the $9 \times 8$ landscape supports \textbf{H4}: \method{}'s relative preference advantage is a joint function of image content and target emotion rather than a uniform constant.

\subsection{Ablation Study}
\label{sec:ablation}

We ablate key components of \method{} on the EmoEdit benchmark using automated metrics (LPIPS, CLIP-I, Emo-S) as a \emph{necessary-but-not-sufficient sanity check}: although Emo-S has an emotion-conditional blind spot to non-stereotypical edits (Sec.~{4.4}), the blind spot is direction-symmetric within our pipeline---it penalises Full \method{} and each ablated configuration alike, so within-run rankings between variants remain trustworthy even though absolute levels are depressed relative to human preference.
Small-scale human-rated ablation is left to future work.
Five ablation configurations and their automated-metric scores are reported in Tab.~{9}.

\begin{table}[t]
\centering
\caption{Ablation study on the EmoEdit benchmark. \emph{w/o Open Discovery}---restricting the planner to a fixed factor taxonomy---causes the largest drop in Emo-S ($0.328 \to 0.224$, $-32\%$). Scores are from a separate inference run intended for \emph{within-run comparison} among ablation variants (Qwen-Image-Edit is non-deterministic under our setup). Best per column in \textbf{bold}---note that higher CLIP-I and lower LPIPS do not necessarily indicate better emotional editing, since conservative outputs stay closer to the source while failing to convey the target emotion (cf.\ Sec.~{4.4}).}
\label{tab:ablation}
\resizebox{\linewidth}{!}{
\begin{tabular}{@{}p{0.62\linewidth}ccc@{}}
\toprule
\textbf{Configuration} & LPIPS$\downarrow$ & CLIP-I$\uparrow$ & Emo-S$\uparrow$ \\
\midrule
  \textbf{Full \method{}} & 0.352 & 0.803 & \textbf{0.328} \\
\midrule
  \textbf{A1: w/o Open Discovery} — planner restricted to fixed factor taxonomy~\cite{yang2024emoedit} & 0.358 & 0.829 & 0.224 \\
  \textbf{A2: w/o Planner Agent} — bare prompt + verifier loops only & 0.468 & 0.787 & 0.277 \\
  \textbf{A3: w/o Plan Verifier} — first plan executed without audit & 0.451 & 0.78 & 0.258 \\
  \textbf{A4: w/o Result Verifier} — first edit accepted as-is & 0.445 & 0.785 & 0.298 \\
  \textbf{A5: Editor only} — no agent, single Qwen pass (same-editor control) & \textbf{0.285} & \textbf{0.846} & 0.213 \\
\bottomrule
\end{tabular}}
\end{table}

\noindent\textbf{Reading.}
Tab.~{9} surfaces a clean ordering of contributions, made unambiguous by the addition of the new \emph{Editor-only} (raw Qwen, no agent) row.
\emph{Editor-only sits at the floor of the editor--planner trade-off}: the strongest LPIPS ($0.285$) and CLIP-I ($0.846$) of any configuration but also the lowest Emo-S ($0.213$, a $-35\%$ drop from Full \method{}).
That is, raw Qwen on the bare prompt ``Make this image evoke [emotion]'' produces edits that are highly faithful to the source but largely fail to convey the target emotion---the system simply does not know \emph{what} to edit, only that something should change.
Adding back \emph{any} planning component lifts Emo-S substantially while accepting some loss of source preservation: \emph{w/o Open Discovery} ($+0.011$ Emo-S) restricts the planner to a fixed factor taxonomy, \emph{w/o Planner Agent (verifiers retained)} ($+0.064$) relies only on the verifier loops to coerce the editor, and the verifier-only ablations sit between these poles.
The largest single jump is from \emph{w/o Open Discovery} to \emph{Full} ($0.224 \to 0.328$, $+47\%$): switching the planner from a fixed taxonomy to open-ended discovery is the most consequential design choice in the pipeline, larger than the contribution of either verifier loop ($0.328 - 0.298 = +0.030$ for the result verifier; $0.328 - 0.258 = +0.070$ for the plan verifier).
\emph{Full \method{} accepts a $-0.061$ CLIP-I cost} (relative to Editor-only) \emph{in exchange for a $+0.115$ Emo-S gain}: the trade-off is itself the signature of open-ended discovery doing its job---semantically meaningful changes that move the image away from the source rather than cosmetically similar ones that do not change the perceived emotion.

\noindent\textbf{Editor difficulty across emotions.}
Across $\nTotalTraces{}$ main-pipeline edit traces, we additionally observe that the edit-refine loop's difficulty varies sharply with the target emotion: the editor hits the maximum iteration cap without verifier approval $\diffRatio{}$ more often for the hardest emotion (\hardestEmo{}, $\hardestEmoRej{}$) than for the easiest (\easiestEmo{}, $\easiestEmoRej{}$).
This per-emotion difficulty ranking converges with the Q4 attribution distribution (Sec.~{4.3}) and the per-emotion correlation analysis (Sec.~{4.4}): emotions that require narrative-element insertion (amusement, fear, anger) are systematically harder for the editor; emotions realised through atmospheric tweaks (contentment, sadness) are easier.
The hardest cases concentrate in a small set of $\nHardestImages{}$ source images that repeatedly hit the edit-refine limit across multiple target emotions, typically because their framing leaves little room for narrative insertions without overwriting the anchor.

The per-component drops in Tab.~{9} show that no single component is dispensable, supporting \textbf{H5}.

\subsection{Qualitative Results: Cross-Source Stability}
\label{sec:qualitative}

Beyond aggregate metrics, we briefly probe the \emph{stability of the same method across heterogeneous sources}.
Our three datasets span three image regimes---curated stock-style photographs (EmoEdit), in-the-wild snapshots (EmoEditor), and stylistically-marked paintings (ArtEmis); a method dependent on training-distribution aesthetics would only succeed on one.
Fig.~{8} shows one source from each regime, each edited toward \emph{Fear}---the same agent, prompts, and hyperparameters across all three.
Even at this single-emotion glance the method (i)~discovers a context-grounded strategy per source (a moonlit night-window with an ominous silhouette outside the interior, a misty forest with a glowing-eyed snarling wolf, and Van Gogh's \emph{Starry Night} re-coloured into purple--green discord with watching eyes in the swirls); (ii)~\emph{re-frames} an already-emotional source (the snarling wolf) by reusing its menacing pose rather than overwriting the subject; and (iii)~preserves the painting's impasto brushstroke style while executing the semantic edit.
Across the full set of Mikels emotions, the same source-dependent pattern persists: \method{} preserves the source anchor while varying atmosphere, narrative elements, or subject behavior according to the target emotion.

\begin{figure}[t]
\centering

\newcommand{\qualW}{0.295\linewidth}

\setlength{\tabcolsep}{2pt}\renewcommand{\arraystretch}{0}%

\begin{tabular}{@{}c@{\hspace{2pt}}ccc@{}}
   &
   {\scriptsize\shortstack{\textbf{EmoEdit}\\(curated photo)}} &
   {\scriptsize\shortstack{\textbf{EmoEditor}\\(snapshot)}} &
   {\scriptsize\shortstack{\textbf{ArtEmis}\\(painting)}} \\[1pt]
   \rotatebox{90}{\scriptsize\textbf{Source}} &
   \includegraphics[width=\qualW]{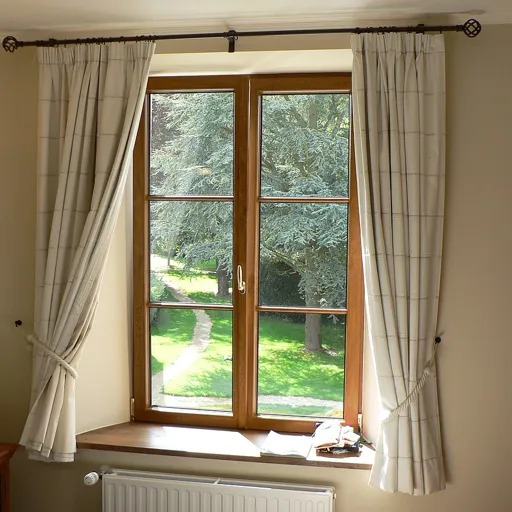} &
   \includegraphics[width=\qualW]{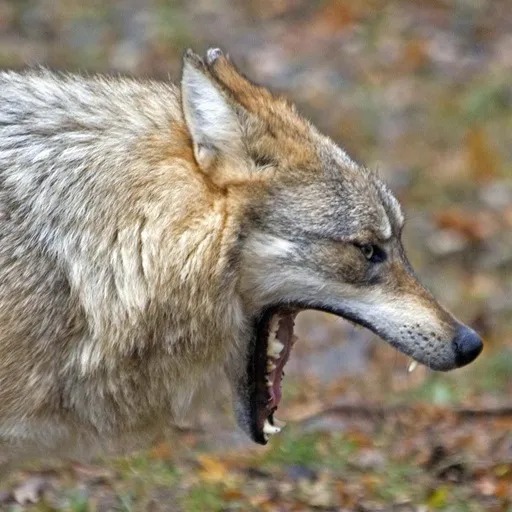} &
   \includegraphics[width=\qualW]{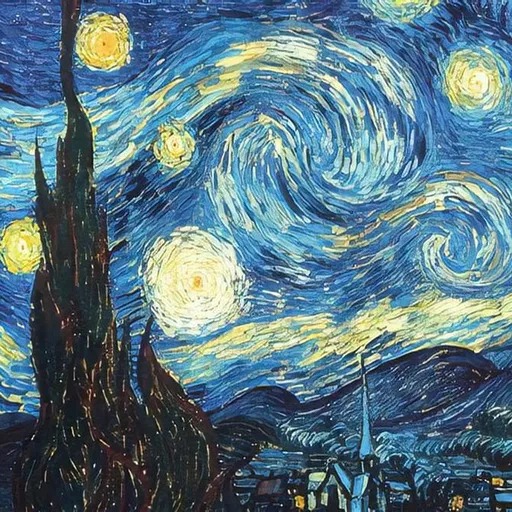} \\[1pt]
   \rotatebox{90}{\scriptsize\textbf{Fear}} &
   \includegraphics[width=\qualW]{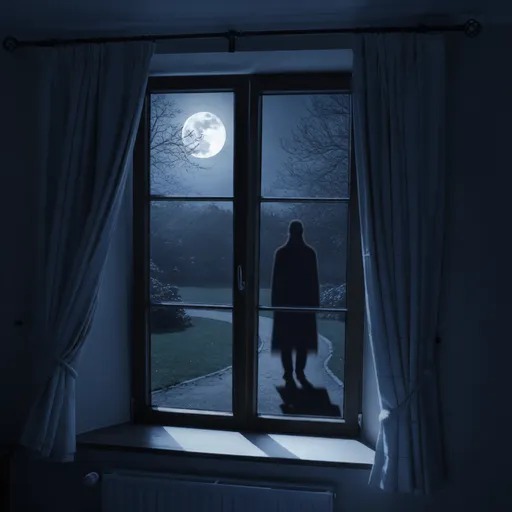} &
   \includegraphics[width=\qualW]{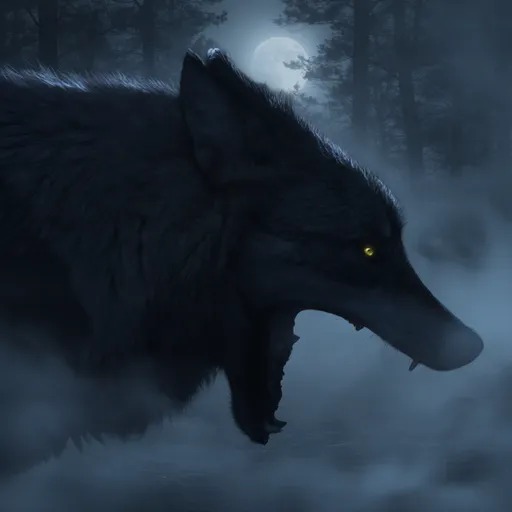} &
   \includegraphics[width=\qualW]{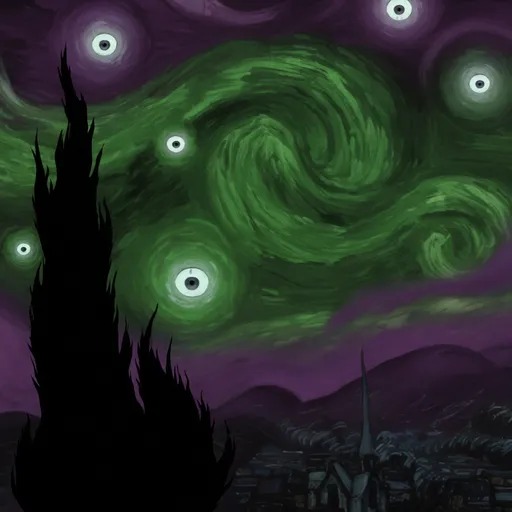} \\
\end{tabular}

    \caption{\textbf{Cross-source stability (Fear only).} Three image regimes (curated photograph, in-the-wild snapshot, iconic painting); per source, \method{} produces a context-grounded \emph{Fear} edit. The examples compactly illustrate the broader cross-source pattern across the Mikels emotion set.}
    \label{fig:qualitative}
\end{figure}

\subsection{Interactive Editing Study}
\label{sec:interactive}

Beyond automated-pipeline quality, emotional editing is also \emph{personal expression}: a real user iterates on intent.
We conduct a small-scale usability pilot with $\nInteractiveParticipants{}$~non-expert participants and $\nInteractiveSessions{}$~plan-review-iterate sessions across the eight Mikels emotions.
Participants review the affordance-level plan, optionally provide natural-language feedback, and then rate the accepted result.
The sample precludes population-level claims; we report this as preliminary deployment evidence.

\noindent\textbf{Behavioural results.}
Every session concluded with an accepted result ($\pctInteractiveAccepted{}$); the workflow converges quickly ($\interactivePlanIterMean{}$~plan and $\interactiveEditIterMean{}$~edit iterations on average, median $\interactiveDurMedian{}$\,s end-to-end); $\pctInteractiveFeedback{}$ of sessions exercised the explicit plan-feedback channel rather than accepting the first plan; mean personal satisfaction reached $\interactiveSatMean{}/5$ with $\pctInteractiveSatHigh{}$ of sessions rated $\geq 4$.

\noindent\textbf{Post-session ratings.}
A post-session 8-item Likert questionnaire from all $\nWebuiRespondents{}$~participants surfaces two findings.
First, the dialog-based plan-feedback feature was endorsed by \emph{every} participant ($\webuiDialogAnyNecessaryPct{}$: $\webuiDialogVeryNecessaryN{}$/$\nWebuiRespondents{}$ ``very necessary,'' the rest ``somewhat necessary''), confirming that affordance-as-interactive-surface is functionally required, not optional polish.
Second, planner-side ratings (initial-plan inspiringness $\webuiQInspiring{}$/5, follow-up understanding $\webuiQUnderstand{}$/5) are uniformly higher than the executed image-quality rating ($\webuiQQuality{}$/5)---a pattern that converges with the Editor-only ablation (Sec.~{4.6}) and with participants' open-ended ``plan is good but image quality is the bottleneck.''
This user-perception evidence independently corroborates that the planner does work the editor cannot.

Together with the controlled pairwise study (Sec.~{4.2}), this pilot provides preliminary evidence for \textbf{H6} (\emph{non-expert users can converge on personalised emotional edits within a small number of plan/edit iterations using \method's interactive workflow}); a larger confirmatory study is left to future work.


\section{Conclusion}
\label{sec:conclusion}

We argued that the solution space of emotional image editing is far larger than existing methods assume, and \method{} addresses this by shifting the paradigm from ``how should I edit?'' to ``what can I edit?''---discovering image-specific strategies through emotion-conditioned reasoning rather than selecting from fixed categories.

Three findings stand out: (i)~humans significantly prefer context-driven edits over fixed-taxonomy ones; (ii)~edit-strategy distribution shifts predictably with target emotion (narrative-heavy for amusement / disgust, atmosphere-heavy for awe / contentment), evidencing \emph{emotion-adaptive} selection; (iii)~classifier-based metrics (Emo-A, Emo-S) exhibit an emotion-conditional measurement blind spot toward non-stereotypical context-grounded edits, varying an order of magnitude across target emotions.
We additionally present the first empirical analysis of a \emph{relative content--emotion preference-affinity landscape}, with practical implications for adaptive intensity selection and stratified benchmarking.

\noindent\textbf{Limitations.}
\label{sec:limitations}
Our framework depends on the VLM's reasoning capabilities for editing potential discovery, and its quality will scale with future VLM improvements.
The iterative dual-loop architecture, while effective, is slower than single-pass methods.
Our user study covers all eight Mikels emotion categories with \nAnswers{} valid responses across \nHumanQuestions{} pairwise questions from \nUniqueParticipants{} unique participants; further scaling participants per emotion would tighten confidence intervals on per-cell measurements within the preference-affinity landscape, and an absolute Likert-scale editability survey would disentangle relative preference from intrinsic difficulty.

\noindent\textbf{Future work.}
Five directions: (1)~formal measures of per-image emotion editability to guide intensity selection; (2)~distilling editing-potential reasoning into an end-to-end model that retains strategy diversity without the iterative cost; (3)~using the released Q4.1 bounding-box annotations as \emph{narrative trigger regions} to test stereotype-vs-semantic-understanding at the spatial level; (4)~a fully editor-controlled head-to-head against agentic AIM systems (EmoAgent, EmoFeedback$^2$, Emotion-Director) on a unified backbone; (5)~per-image hierarchy decomposition (Anchors / Variables / Context) cross-referenced with the preference-affinity landscape.

{\small
\bibliographystyle{ieeenat_fullname}
\bibliography{references}
}

\end{document}